\documentclass[11pt, letterpaper, logo, onecolumn, copyright]{main}

\usepackage[authoryear, sort&compress, round]{natbib}

\usepackage[inkscapeformat=png]{svg}

\usepackage[most, breakable, skins]{tcolorbox}

\tcbuselibrary{skins}
\usepackage{lipsum}
\usepackage{tabularx}
\usepackage{afterpage}
\usepackage{booktabs}
\usepackage{xurl} 
\usepackage{subcaption}
\usepackage{makecell}
\usepackage{multirow}
\usepackage{bm}
\usepackage{multicol}
\usepackage{array}
\usepackage{float}
\usepackage{listings, listings-rust}
\usepackage{fontawesome5}
\usepackage{hyperref}
\usepackage{amssymb}
\usepackage{graphicx}
\usepackage[dvipsnames]{xcolor}
\usepackage{cleveref}
\usepackage{longtable}
\usepackage{pdflscape}
\usepackage{adjustbox}
\usepackage{nicematrix}
\usepackage{CJKutf8}
\usepackage{ragged2e}
\usepackage{colortbl}
\usepackage{enumitem}
\usepackage{bbm}
\usepackage{pifont}
\usepackage{hyphenat}
\usepackage{amsmath}
\usepackage{amsthm}  
\usepackage{circledsteps}
\usepackage{diagbox}
\usepackage{bookmark}
\usepackage{textcomp}

\usepackage{fvextra}

\usepackage{algorithm} 
\usepackage{algpseudocode}

\usepackage{wrapfig}
\usepackage{xspace}
\usepackage{tikz}
\usepackage[normalem]{ulem}

\usepackage{pgfplots}
\usepackage{pgfplotstable}
\pgfplotsset{compat=1.18}

\usepackage[most]{tcolorbox}
\definecolor{lightblue}{RGB}{210, 220, 250}

\definecolor{medgray55}{gray}{0.55}
\definecolor{medgray}{gray}{0.7}
\definecolor{litegray}{gray}{0.9}
\definecolor{gblue}{RGB}{210, 227, 252}
\definecolor{gred}{RGB}{250, 210, 207}
\definecolor{gyellow}{RGB}{254, 239, 195}
\definecolor{ggreen}{RGB}{206, 234, 214}
\definecolor{gorange}{RGB}{254, 223, 200}

\definecolor{gblue9}{RGB}{23, 78, 166}
\definecolor{gred9}{RGB}{165, 14, 14}
\definecolor{gyellow9}{RGB}{227, 116, 0}
\definecolor{ggreen9}{RGB}{13, 101, 45}
\definecolor{gorange9}{RGB}{176, 96, 0}

\definecolor{myblue}{rgb}{0,0,1}
\definecolor{myred}{rgb}{1,0,0}
\definecolor{mylightgray}{gray}{0.95}
\definecolor{myCite}{HTML}{1C4587}

\definecolor{highlightblue}{HTML}{185ABC}
\definecolor{cellHighlight}{HTML}{dbefff}
\definecolor{lightgray}{RGB}{211, 211, 211}
\definecolor{lightfont}{gray}{0.3}

\usepackage{minitoc}

\noptcrule


\newcolumntype{L}[1]{>{\raggedright\let\newline\\\arraybackslash\hspace{0pt}}m{#1}}
\newcolumntype{C}[1]{>{\centering}m{#1}}

\newcolumntype{R}[1]{>{\raggedleft\let\newline\\\arraybackslash\hspace{0pt}}m{#1}}

\definecolor{ao}{rgb}{0.0, 0.0, 1.0}

\newcommand\vcent[1]{\vcenter{\hbox{#1}}}

\newcommand\loudspeaker[1][3]{\ensuremath{\vcent{\rule{.6ex}{.6ex}}\kern-.5ex
  \vcent{\scalebox{.6}[1]{\rotatebox[origin=center]{90}{$\blacktriangle$}}}
  \ifnum#1>0\relax\kern.05ex\vcent{\scalebox{.4}{\ttfamily)}}
  \ifnum#1>1\relax\kern-.4ex\vcent{\scalebox{.56}{\ttfamily)}}
  \ifnum#1>2\relax\kern-.55ex\vcent{\scalebox{.7}{\ttfamily)}}
  \fi\fi\fi}
}

\makeatletter
\renewcommand\subparagraph{
 \@startsection {subparagraph}{5}{\z@ }{3.25ex \@plus 1ex
 \@minus .2ex}{-1em}{\normalfont \normalsize \bfseries }}
\makeatother

\let\cite\citep
\hypersetup{
  citecolor = myCite,  
  linkcolor = myCite,   
  urlcolor  = myCite
}
\newcommand{\myheaderbreak}{\\}
\title{
Beyond Scaling: Measuring and Predicting the Upper Bound of Knowledge Retention in Language Model Pre-Training
}

\author{
  Changhao Jiang\textsuperscript{\rm *\dag},
  Ming Zhang\textsuperscript{\rm *},
  Yifei Cao\textsuperscript{\rm *},\\
  Junjie Ye\textsuperscript{\rm },
  Xiaoran Fan\textsuperscript{\rm },
  Shihan Dou\textsuperscript{\rm },
  Zhiheng Xi\textsuperscript{\rm },
  Jiajun Sun\textsuperscript{\rm },
  Yi Dong\textsuperscript{\rm },
  Yujiong Shen\textsuperscript{\rm },\\
  Jingqi Tong\textsuperscript{\rm },
  Baoyu Fan\textsuperscript{\rm },
  Tao Gui\textsuperscript{\rm },
  Qi Zhang\textsuperscript{\rm \dag},
  Xuanjing Huang\textsuperscript{\rm }\\
  \vspace{0.3cm} 
  \normalsize 
  Fudan NLP Group, IEIT Systems Co.\ Ltd. \\
  \texttt{\normalsize chjiang25@m.fudan.edu.cn, qz@fudan.edu.cn}
}

\vspace{-50pt}

\begin{abstract}
The GPT-4 technical report suggests that downstream performance can be predicted from pre-training signals, but offers little methodological detail on how to quantify this. This work address this gap by modeling knowledge retention, the capacity of a pre-trained language model to memorize factual information from its corpus, and introduce a principled method to estimate it prior to training. We propose Size-dependent Mutual Information (SMI), an information-theoretic predictor that integrates knowledge frequency, knowledge specificity, and model size to forecast closed-book question answering (QA) accuracy. SMI is validated through large-scale document retrieval over the disclosed pre-training corpora of 21 public and 3 custom models, combined with a robust multi-template QA evaluation. Experiments show that SMI significantly outperforms repetition-based baselines and achieves R² > 0.7 in predicting QA accuracy for models above 1B parameters, without additional training. The analysis further reveals diminishing returns from scaling data and model size and provides evidence for an intrinsic upper bound on knowledge retention achievable by pre-training alone, motivating retrieval and other augmentation strategies.
The dataset and code are available at \url{https://github.com/yuhui1038/SMI}.
\end{abstract}

\begin{document}
\doparttoc
\faketableofcontents
\begingroup
  \renewcommand\thefootnote{}
  \footnote{\hspace{-1.8em}\textsuperscript{*}Equal Contribution.\\
            \textsuperscript{\dag}Corresponding Author.}
\endgroup
\vspace{-30pt}
\maketitle
\renewcommand{\myheaderbreak}{ } 
\vspace{-15pt}

\begin{figure*}[tbh]
\begin{center}
\includegraphics[width=\textwidth]{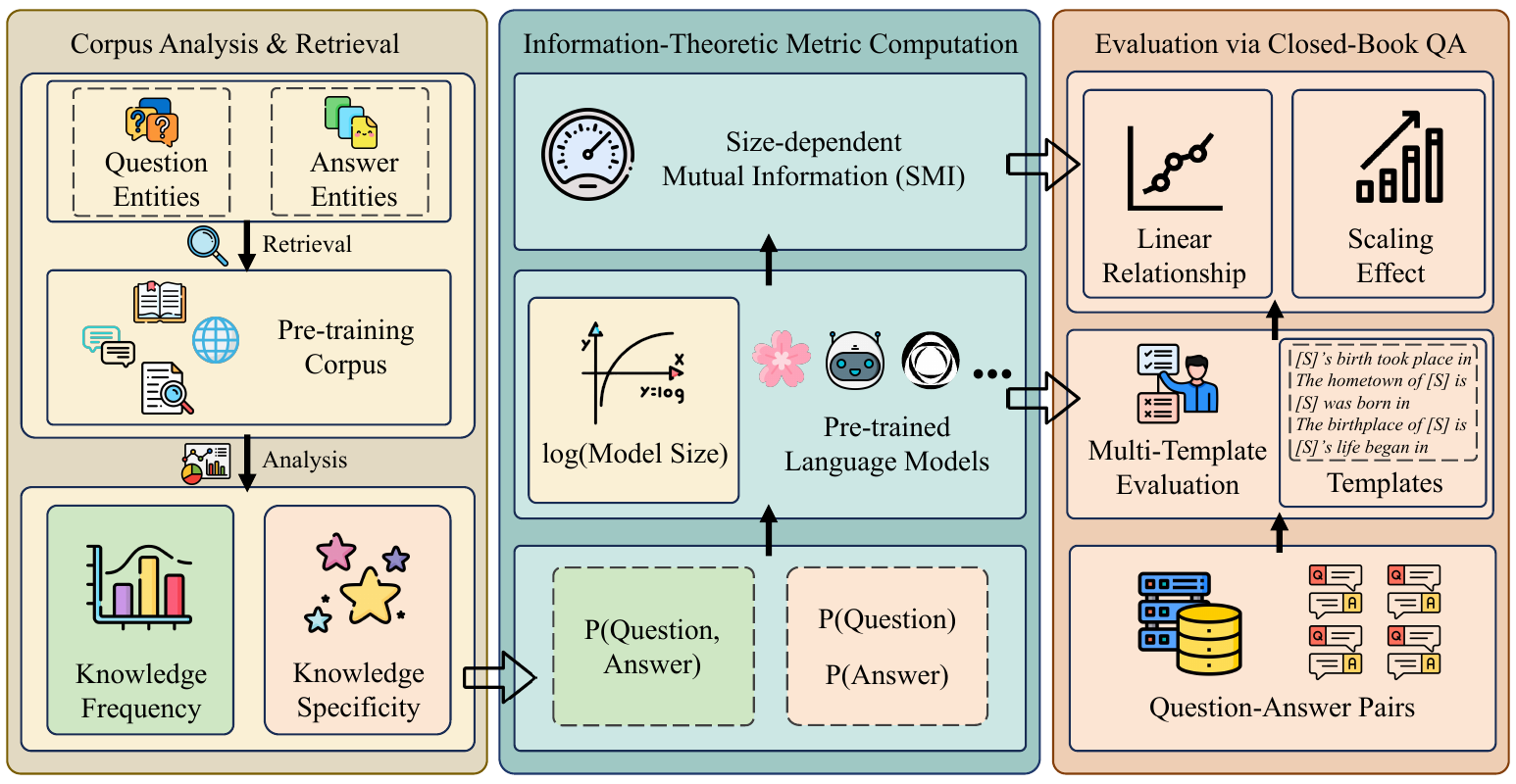}
\caption{Overview of our method. The framework consists of three stages: (1) Corpus Analysis and Retrieval, where subject and object (question and answer) entities are extracted from pre-training data and their frequencies and co-occurrences are computed; (2) Information-Theoretic Metric Computation, where knowledge frequency, knowledge specificity, and model size are integrated into the size-dependent mutual information (SMI) metric; and (3) Evaluation via Closed-Book QA, where pre-trained language models are assessed using multi-template evaluation, and SMI is compared with empirical QA accuracy to analyze linear relationships and scaling effects.}
\label{intro}
\end{center}
\end{figure*}

\section{Introduction}
\label{introduction}

The technical report on GPT-4 states, "We registered predictions for GPT-4's performance on HumanEval before training completed, using only information available prior to training"~\citep{DBLP:journals/corr/abs-2303-08774}. Understanding and predicting the generalization ability of large language models (LLMs) has become a key research question in the study of scaling laws~\citep{DBLP:journals/corr/abs-2001-08361}. Although prior work has established empirical relationships between loss, data size, and model parameters, these results primarily offer coarse-grained predictions of overall performance and do not address what specific factual knowledge is retained from pre-training corpora. Recent studies have shown that LLMs can memorize and regurgitate training data~\citep{carlini2021extracting, biderman2023emergent}, but these findings are largely descriptive and do not provide a systematic way to predict retention for new model–corpus pairs. Developing principled approaches for this purpose is crucial both for improving our theoretical understanding of model training dynamics and for guiding practical decisions on data curation and compute allocation~\citep{hoffmann2022training, alabdulmohsin2022revisiting, ruan2405observational}.

This work addresses this gap by modeling knowledge retention, the capacity of a pre-trained language model to memorize factual information from its corpus, and introduces a principled method to reliably estimate it prior to training. Rather than focusing only on aggregate loss curves or perplexity, we take a more fine-grained and systematic view: we ask, given a particular fact in the pre-training corpus, what is the probability that a model will be able to accurately reproduce it in a closed-book setting after training~\citep{DBLP:conf/emnlp/PetroniRRLBWM19}? We term this capacity knowledge retention, and formalize it as a quantitative property of a model–corpus pair. This framing allows us to reason about model memory at the level of individual facts and categories, and to make meaningful predictions before committing to the cost of full-scale training.

We propose Size-dependent Mutual Information (SMI), an information-theoretic predictor that integrates knowledge frequency, knowledge specificity, and model size to forecast closed-book question answering (QA) accuracy~\citep{DBLP:conf/emnlp/RobertsRS20}. Since pre-trained language models are optimized only for next-token prediction rather than explicit knowledge retrieval, closed-book QA provides a natural and challenging probe for assessing factual recall. Our formulation builds on the intuition that both the frequency of a fact in the corpus and its rarity influence its likelihood of being learned, and that these effects are jointly influenced by model size. Concretely, SMI captures the mutual information between knowledge subjects and objects (or questions and answers), combining these statistics with model size. The resulting predictor is simple to compute from pre-training data and scales gracefully with corpus size, making it suitable for practical use in large-scale settings.

SMI is validated through large-scale document retrieval over the disclosed pre-training corpora of 21 public and 3 custom models, combined with a robust multi-template QA evaluation. Our experimental setup reconstructs the exposure of facts in the training data, then evaluates whether models can correctly answer corresponding questions in a closed-book setting. We design multi-template QA prompts to reduce prompt variance and report results across a diverse set of model families, covering a range from 14M to 176B parameters.

Experiments show that SMI significantly outperforms repetition-based baselines and achieves the coefficient of determination (R²) > 0.7 in predicting QA accuracy for models above 1B parameters, without additional training. Relative to heuristics based solely on repetition counts, SMI yields a stronger correlation with observed QA accuracy, particularly in the large-model regime where memorization behavior appears more stable. Moreover, we observe that SMI maintains comparable predictive performance across multiple model families and domains, indicating that it captures a generalizable aspect of factual knowledge acquisition in pre-trained language models.

The analysis further reveals diminishing returns from scaling both data and model size, suggesting that pre-training alone faces an intrinsic limit in its ability to retain factual knowledge. Recent studies have similarly observed a performance ceiling on knowledge-intensive tasks, even with substantial scaling of parameters and data~\citep{zhang2025llmeval}. These observations point to inherent limitations of purely parametric approaches: beyond a certain scale, further increases in model size or training data yield only marginal gains in closed-book performance. This motivates exploring hybrid approaches that augment parametric memory with non-parametric mechanisms, including retrieval-augmented generation, external memory modules, or targeted fine-tuning, to mitigate the ceiling effects observed in pre-training.
Overall, our contributions are threefold:
\begin{enumerate}
    \item We formalize knowledge retention as a quantitative metric defined over model–corpus pairs, providing a new way to analyze what models actually memorize during pre-training.
    \item We propose SMI, a data-driven and training-free predictor that integrates knowledge frequency, knowledge specificity, and model size to estimate closed-book QA accuracy.
    \item We empirically evaluate SMI on 24 models, showing that it achieves strong predictive accuracy, characterizing diminishing returns from scaling data and model size, and suggesting the presence of an upper bound on knowledge retention achievable by pre-training alone.
\end{enumerate}

\section{Related Work}
\label{relatedwork}

\paragraph{Pre-training data and knowledge retention}
The characteristics of pre-training data are crucial for knowledge retention and downstream performance in LLMs. Many studies emphasize the role of repetition frequency. For example, \citet{DBLP:conf/iclr/CarliniIJLTZ23} identified a logarithmic relationship between repetition and memory effects, while \citet{DBLP:journals/jmlr/ChowdheryNDBMRBCSGSSTMRBTSPRDHPBAI23} showed that sequences repeated over 500 times are completed with over 40\% accuracy. \citet{DBLP:conf/acl/JuCY0DZL24} found that high-frequency data can lead to ``factual shortcuts'' in multi-hop reasoning. \citet{DBLP:journals/corr/abs-2404-05405} suggested that exposing knowledge 1000 times enables a storage capacity of two bits per parameter. Additionally, \citet{DBLP:conf/emnlp/RazeghiL0022} and \citet{DBLP:journals/corr/abs-2311-00871} linked lower-order co-occurrences to improved numerical reasoning, while \citet{DBLP:journals/corr/abs-2309-13638} associated task prevalence in pre-training with better performance on tasks like ROT13.
Furthermore, \citet{DBLP:conf/nips/ChangPYYSCS24} revealed that simply duplicating data might not inherently enhance knowledge retention, underscoring the importance of data diversity over mere repetition.

\paragraph{Model size and knowledge retention}
Furthermore, the model size significantly influences the knowledge retention of LLMs. Research by OpenAI demonstrates that as the number of non-embedding parameters in a model increases, the test loss decreases following a power law pattern~\citep{DBLP:journals/corr/abs-2001-08361}.
~\citet{DBLP:conf/icml/KandpalDRWR23} evaluate the BLOOM model families~\citep{DBLP:journals/corr/abs-2211-05100} on long-tail QA tasks and find that the fact-based QA accuracy is highly linearly correlated with the logarithm of the number of model parameters. Additionally, the memory capacity of a model closely correlates with its size~\citep{DBLP:journals/corr/abs-2403-00510}.

\paragraph{Knowledge triples as an evaluation tool}
Knowledge triples (subject, relation, object) serve as an effective tool for evaluating the knowledge retention of LLMs.~\citet{DBLP:conf/emnlp/PetroniRRLBWM19} employ the LAMA method to evaluate the implicit knowledge of models like BERT, demonstrating the value of knowledge triples in reasoning tasks. Subsequently,~\citet{DBLP:journals/corr/abs-2402-14273, DBLP:conf/acl/JuCY0DZL24} further explore the application of knowledge triples in reasoning tasks. Moreover,~\citet{DBLP:conf/emnlp/WangYXQD00GJX0C24} and~\citet{DBLP:conf/icml/Allen-ZhuL24} use knowledge triples to explore LLMs' ability to retain and extract knowledge.

\section{Methodology}
\label{methodology}
To predict a model’s ability to recall factual knowledge before training, we develop a framework that estimates closed-book QA accuracy directly from pre-training data statistics and model size. We first revisit prior work and evaluate a simple repetition-based co-occurrence metric as a baseline. Building on information-theoretic principles, we then introduce the Mutual Information (MI) metric, which accounts for both the frequency and specificity of knowledge items and yields improved predictive performance. Finally, we extend MI to SMI, which incorporates model size as a scaling factor, providing a more accurate and scalable predictor of knowledge retention.

\subsection{Starting from the Co-occurrence Metric}
\paragraph{Knowledge triples}
In QA tasks, entity relationships can be represented as knowledge triples, denoted by tuples $t = (s, r, o)$, where $s$ is the subject, $r$ is the relation, and $o$ is the object~\citep{DBLP:conf/emnlp/WangYXQD00GJX0C24}. For a given LLM, denoted as $F$, we define $F$ as mastering the knowledge triple $t$, thus demonstrating competence in the corresponding QA task, as follows:
\begin{equation}
    F(q_{s,r}) = o, \quad (q \in \mathbf{Q}, o \in \mathbf{O}).
\end{equation}

Here, \(q_{s,r}\) represents the combined representation of the subject \(s\) and the relation \(r\), while \(o\) denotes the representation of the object. For instance, if \(s = \text{Apple}\) and \(r = \text{headquarter}\), then \(q_{s,r}\) can be expressed as statements like ``The headquarters of Apple is in...'', ``Apple is headquartered in...'', or ``Apple's head office is based in...''. Similarly, \(o\) corresponds to object representation, such as ``Cupertino'', ``Cupertino, California'', or ``Cupertino city''. The sum of all \(q_{s,r}\) representations is denoted as \(\mathbf{Q}\), and the sum of all \(o\) representations is denoted as \(\mathbf{O}\).

\paragraph{Co-occurrence metric baseline}
Prior work indicates that the more frequently knowledge appears in pre-training data, the better the model learns~\citep{DBLP:conf/icml/KandpalDRWR23, DBLP:conf/acl/MallenAZDKH23, DBLP:journals/corr/abs-2404-05405}. Specifically, linking the entities in the questions and answers of a QA task to the pre-training data and counting the number of documents that jointly link the question and answer entities is a feasible approach~\citep{DBLP:conf/icml/KandpalDRWR23}. There is a linear relationship between QA accuracy and the logarithm of the number of documents~\citep{DBLP:conf/iclr/CarliniIJLTZ23, DBLP:conf/icml/KandpalDRWR23}. 

To quantify such co-occurrence effects, let \(\mathbf{N}\) denote the total number of documents in the pre-training corpus. We define \(P(s)\) as the proportion of documents containing the subject entity \(s\), \(P(o)\) as the proportion of documents containing the object entity \(o\), and \(P(s, o)\) as the proportion of documents containing both \(s\) and \(o\).
Based on this formulation, we define the co-occurrence metric as \(\log(N \cdot P(s, o))\), which corresponds to the logarithm of the number of documents in which the \(s\) and \(o\) co-occur. We use this metric as a baseline for predicting closed-book QA accuracy.

\paragraph{Multi-template evaluation method}
Evaluating the knowledge retention of LLMs presents challenges due to the diverse range of queries formed by the subject and relation of knowledge triples. Inspired by the evaluation method for fine-tuned LLMs~\citep{DBLP:journals/corr/abs-2409-15825}, we develop a robust multi-template evaluation method to assess the accuracy of pre-trained models in QA tasks.

Specifically, by constructing synonymous sentences, we generate a substantial number of query templates that are semantically similar but vary in form. For each type of knowledge triple, we produce 20 templates that differ in length and structure. Each template $q$ represents a specific instance within the query set \(\mathbf{Q}\), and these 20 selected $q$ templates approximate the entirety of \(\mathbf{Q}\).

\paragraph{Linear relationship}
On the one hand, We perform document-level retrieval on the pre-training data and calculate the co-occurrence metric. For consistency with information-theoretic conventions \citep{DBLP:journals/bstj/Shannon48a}, all logarithms in this paper use base 2. Since specifying the subject and object typically determines the relation, we omit the relation component for simplicity.

On the other hand, by employing the multi-template evaluation approach, we measure the model’s accuracy on QA tasks. To examine the relationship between the co-occurrence metric and QA accuracy, we fit a linear regression model and compute the R² along with the Mean Squared Error (MSE). These metrics allow us to evaluate both the strength and the stability of the linear relationship.

\subsection{Considering the Knowledge Specificity to Obtain the MI Metric}

\paragraph{Knowledge specificity}
The co-occurrence metric does not account for the specificity of knowledge. For a knowledge triple $ t = (s, r, o) $, when the subject entity $ s $ and the object entity $ o $ appear frequently in the pre-training data, corresponding to higher values of $ P(s) $ and $ P(o) $, the triple tends to be less specific. In such cases, the association between $ s $ and $ o $ is less distinctive, which makes it more difficult for the model to reliably retain the corresponding factual knowledge.
Conversely, when $ s $ and $ o $ appear infrequently in the pre-training data, corresponding to lower values of $ P(s) $ and $ P(o) $, the knowledge triple is more specific, which allows the model to more easily retain the corresponding fact.
For example, the address of NVIDIA's headquarters is easier to remember than that of Apple's headquarters, because the word "Apple" occurs more frequently across diverse contexts. MI in information theory considers both the co-occurrence of two variables and their individual specificities~\citep{DBLP:journals/bstj/Shannon48}. Therefore, we aim to use MI to improve the co-occurrence metric.

\paragraph{MI metric}
Unlike the MI in information theory, which is defined over the full joint distribution of random variables and accounts for both occurrence and non-occurrence events, the MI metric is computed only over observed occurrences of the subject and object in the corpus.
Given pre-training data \(\mathbf{P}\) consisting of \(\mathbf{N}\) paragraphs, a knowledge triple \((s, r, o)\), and corresponding \(P(s)\), \(P(o)\), and \(P(s, o)\), the MI metric between \(s\) and \(o\) is defined as:
\begin{align}
    I(s, o) &= P(s, o) \log \left( \frac{P(s, o)}{P(s)P(o)} \right), \\
    MI(s, o) &= Norm(\log(I(s, o))).
\end{align}

\begin{figure*}[tb]
\centering
\begin{minipage}{0.8\linewidth}
    \centering
    \includegraphics[width=0.45\linewidth]{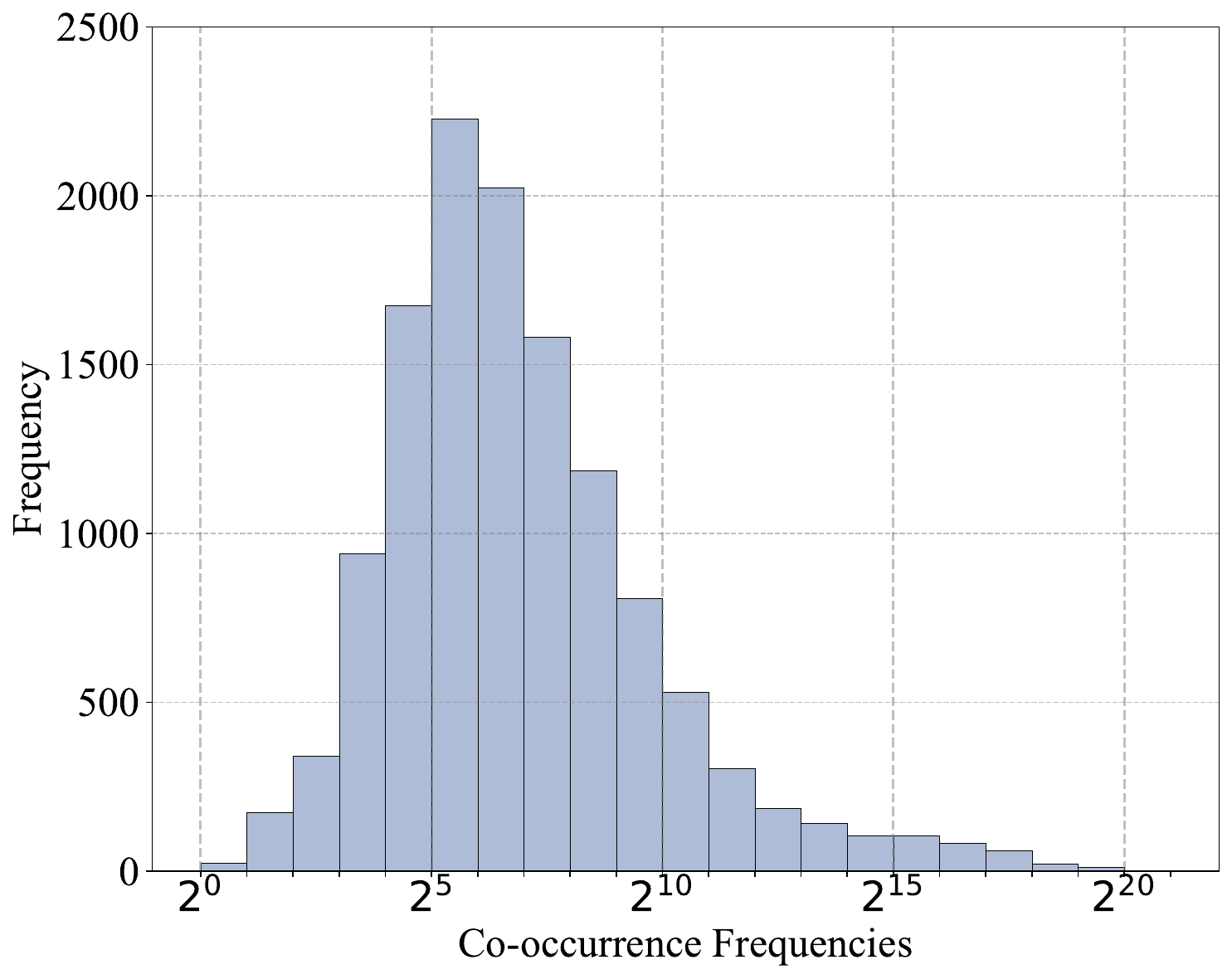}
    \hfill
    \includegraphics[width=0.45\linewidth]{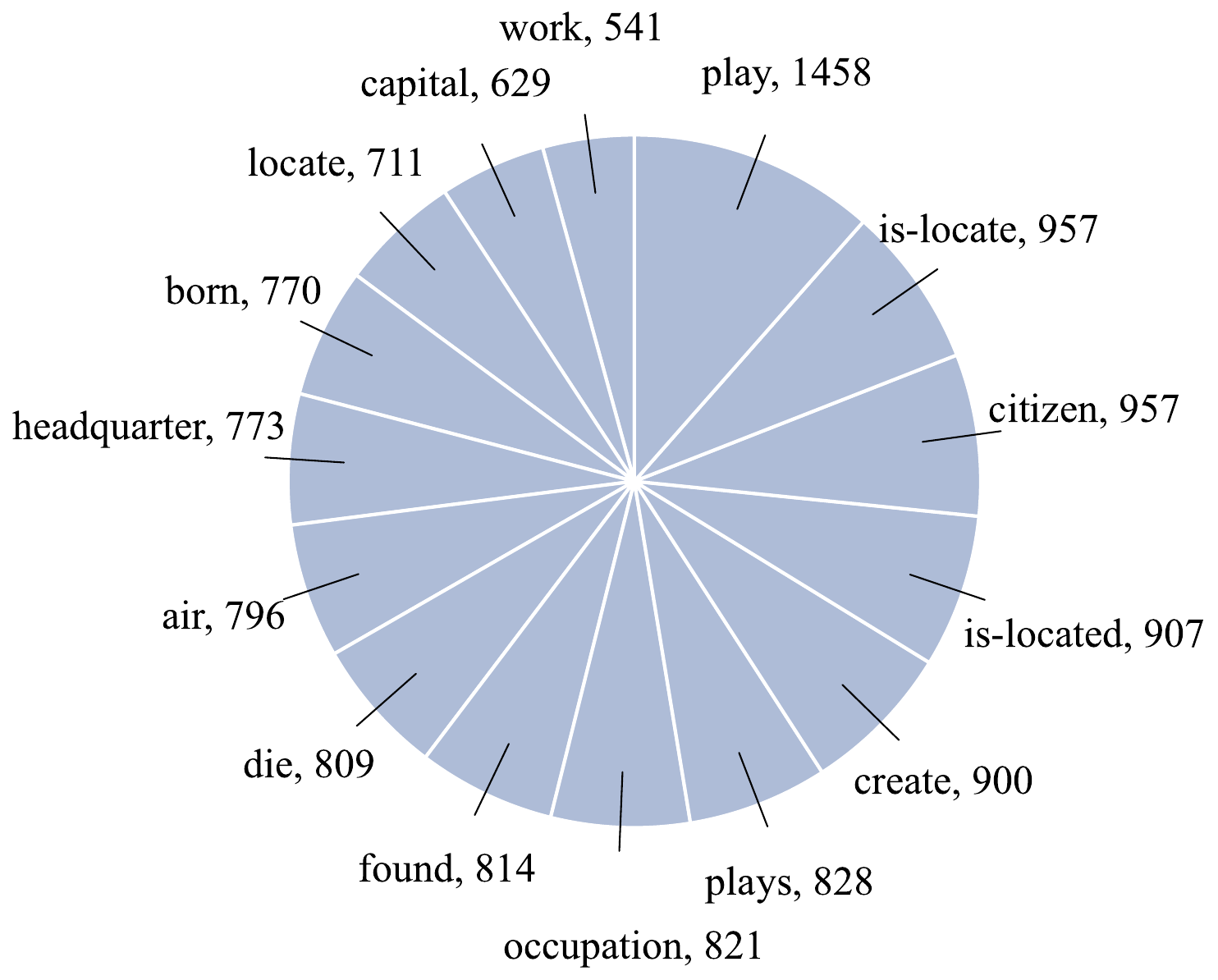}
    \vspace{-4pt}
    \caption{Evaluation data statistics. Left: Co-occurrence frequencies of QA entity pairs in The Pile corpus. Right: Distribution of 12,671 QA tasks across 15 relation types.}
    \label{eval_data}
\end{minipage}
\end{figure*}

In the above formula, \(P(s)\) and \(P(o)\) penalize the number of documents containing the subject and object, respectively, to emphasize the specificity of \(s\) and \(o\). The term \(I(s, o)\) represents the amount of shared information between \(s\) and \(o\), and it can also indicate the reduction in the amount of information about \(o\) given \(s\). During inference with LLMs, this amount of information refers to the probability of the model generating the object given the subject. Considering the uneven distribution of \(I(s, o)\) (see Figure~\ref{eval_data}), we apply a logarithmic transformation followed by normalization to map its values to the range $[0, 1]$, resulting in the MI metric. In rare cases where \(I(s, o)\) becomes zero or negative due to data sparsity or noise, such instances are regarded as anomalies and excluded from further analysis.

\subsection{Considering the Model Size to Obtain the Size-dependent MI metric}

\paragraph{Model size}
The MI metric does not account for the impact of model size. Existing research indicates that the capabilities of LLMs increase with model size, but the returns diminish as the size becomes larger~\citep{DBLP:journals/corr/abs-2001-08361}. In long-tail QA tasks, the QA accuracy is highly linearly correlated with the logarithm of the number of model parameters~\citep{DBLP:conf/icml/KandpalDRWR23}. Therefore, we propose to extend the MI metric by explicitly incorporating the parameter count \(\Phi\) as a measure of model size.

The method of integrating model size is not unique. A guiding principle is that, for the same data, a higher \(\Phi\) should correspond to a higher value of the new metric, reflecting improved QA performance. However, according to scaling laws~\citep{DBLP:journals/corr/abs-2001-08361}, increasing model size exhibits diminishing marginal returns; therefore, the model scale is expressed in logarithmic form. Importantly, as \(\Phi\) approaches infinity, the new metric should converge to the MI metric, indicating that the model’s memory capacity is sufficient to fully capture the relevant knowledge.

\paragraph{Size-dependent MI metric}
Since linear transformations do not alter the strength of linear relationships, an exponential formulation offers a simple yet effective way to incorporate model size while capturing non-linear effects. Based on these principles, we propose the SMI metric, an exponential function with the MI metric as the base and $1 + \frac{1}{\log \Phi}$ as the exponent.
\begin{equation}
    SMI(s, o, \Phi) = MI(s, o)^{1 + \frac{1}{\log\Phi}}.
\end{equation}

In the above formula, the MI metric quantifies the information of knowledge triples in the pre-training data, while the model size \(\Phi\) implies the memory capacity of LLMs. These two factors jointly determine the model's knowledge retention. Since the MI metric is always less than 1, the factor \(1 + \frac{1}{\Phi}\) decreases as \(\Phi\) increases, leading to a larger SMI value, which indicates stronger knowledge retention.

\section{Experiments}
\label{experiments}

In the Experimental Setup, we first curate and enhance an open-source Wikipedia knowledge triple dataset to make it suitable for inference with pre-trained models. We then introduce five large-scale open-source pre-training corpora and perform retrieval-based analysis on them, followed by a detailed description of five model families and our evaluation protocols. In the Experimental Results, we evaluate 21 open-source and 3 in-house pre-trained models, providing a comparative analysis across metrics.

\subsection{Experimental Setup}
\paragraph{Evaluation data} 
We make use of the high-quality knowledge triple dataset ParaRel, which is designed to evaluate whether language models align with factual knowledge, a task similar to ours~\citep{DBLP:journals/tacl/ElazarKRRHSG21}. We select 15 categories of knowledge triple relations with a larger quantity of data and clearer semantics, totaling 12,671 knowledge triples, as shown in Figure~\ref{eval_data}. For each category of knowledge triple relations, we create 20 synonymous query templates to construct the QA task evaluation dataset.

\paragraph{Pre-training corpora} 
We conduct a document-level retrieval analysis of the following pre-training datasets, selected based on their usage in the LLMs under our consideration:
\begin{itemize}[leftmargin=*,itemsep=1mm]
    \item \textbf{The Pile:} This is an open-source, large-scale text dataset specifically designed for training LLMs. It encompasses a variety of sources, including webpages, academic papers, code, and books, with an emphasis on diversity and high quality~\citep{DBLP:journals/corr/abs-2101-00027}.
    \item \textbf{ROOTS (En):} This dataset is the English subset of ROOTS, a comprehensive multilingual corpus constructed by the BIGScience project. It is used for training the BLOOM models and emphasizes linguistic diversity and data transparency~\citep{DBLP:conf/nips/LaurenconSWAMSW22, DBLP:journals/corr/abs-2211-05100}.
    \item \textbf{SlimPajama:} Constructed by Cerebras, this open-source dataset is a large-scale text corpus aimed at providing high-quality training materials for LLMs~\citep{cerebras2023slimpajama}.
    \item \textbf{RefinedWeb:} Developed by the Abu Dhabi Technology Innovation Institute (TII), this large-scale English web dataset is based on CommonCrawl and undergoes rigorous content extraction, filtering, and deduplication processes. It is intended to serve as high-quality pre-training material for LLMs such as Falcon LLM~\citep{DBLP:conf/nips/PenedoMHCACPAL23}.
    \item \textbf{Wikipedia:} This structured text corpus contains Wikipedia articles and is extensively used for language model training and evaluation due to its high-quality content, comprehensive coverage, and semantic consistency~\citep{wikidump}.
\end{itemize}

\paragraph{Pre-training data retrieval} 
We process the pre-training corpus at the document level. For each knowledge triple, we retrieve the frequency of occurrences of the subject, the object, and their co-occurrences. Using the retrieval results from all documents, we calculate the co-occurrence metric and MI metric for each knowledge triple. Subsequently, we compute the SMI metric based on the MI metric and the model size. The retrieval process runs on 20 CPUs for 15 days.

\paragraph{Pre-trained models} 
We conduct experiments on 21 fully open-source pre-trained Transformer decoder-only pre-trained LLMs~\citep{DBLP:conf/nips/VaswaniSPUJGKP17}. These models not only make the model weights available but also provide complete pre-training data. They are primarily categorized into the following families:
\begin{itemize}[leftmargin=*,itemsep=1mm]
    \item \textbf{BLOOM:} Created by BigScience, these open-source multilingual LLMs are trained on ROOTS and supports 46 natural languages and 13 programming languages, with parameter sizes ranging from 560 million to 176 billion~\citep{DBLP:journals/corr/abs-2211-05100}.
    \item \textbf{GPT-Neo:} This family includes the GPT-Neo, GPT-J, and GPT-NeoX models, ranging from 125 million to 20 billion parameters, all trained on The Pile~\citep{gpt-neo, DBLP:journals/corr/abs-2204-06745, mesh-transformer-jax}.
    \item \textbf{Pythia:} Developed by EleutherAI, this fully open-source series of LLMs ranges from 14 million to 12 billion parameters. These models use a uniform data and training process, facilitating research that is both replicable and interpretable~\citep{DBLP:conf/icml/BidermanSABOHKP23}.
    \item \textbf{TinyLlama:} Originating from the Singapore University of Technology and Design (SUTD), this open-source LLM is trained on SlimPajama and Starcoderdata. It has 1.1 billion parameters and is designed with the Llama 2 architecture and tokenizer to provide efficient and lightweight LLM solutions~\citep{DBLP:journals/corr/abs-2401-02385, cerebras2023slimpajama, DBLP:journals/tmlr/LiAZMKMMALCLZZW23}. The specific model we utilize is the TinyLlama-1.1B-intermediate-step-480k-1T, which is trained for approximately one epoch.
\end{itemize}

Additionally, we pre-train three models, named ours-1.6b, ours-7b, and ours-13b, based on RefinedWeb and Wikipedia as a supplement to the aforementioned models.

\paragraph{Evaluation setting} 
We evaluate these pre-trained models on the designated QA evaluation dataset. During inference, we configure the generation parameters with a temperature of 0.7, a maximum of 32 new tokens, and bfloat16 precision. A model response is considered correct if the expected object entity appears anywhere within the generated text. To obtain stable and reliable estimates of performance, we repeat the inference procedure 20 times for each of the 20 queries in the QA task, resulting in a total of 400 responses per model. The final accuracy is computed as the mean over all 400 responses. The full evaluation runs for approximately five days on 8 H100 GPUs.

The retrieval results from the pre-training corpus show that the co-occurrence frequency distribution of question and answer entities is highly skewed (see Figure~\ref{eval_data}). Most QA pairs co-occur between 10 and 1000 times, while only a small portion appears at very low or very high frequencies. To mitigate this imbalance and enable meaningful evaluation across different frequency ranges, we group QA data with similar metric values. For the co-occurrence and MI metrics, we compute average accuracy by binning values into intervals of 0.2 on the unnormalized scale. The SMI metric is then derived from the MI metric after binning.

Finally, we perform linear regression on the metrics and accuracy, and calculate the R² to assess the strength of the linear relationship. Simultaneously, to evaluate the prediction error of the regression equation in specific QA tasks, we compute the MSE for each QA task and take the average.

\begin{table*}[ht]
    \caption{R² and MSE ($\times 10^{-2}$) results of the CO-OCCUR, MI, and SMI metrics across different models.}
    \label{main_table}
    \begin{center}
    \resizebox{0.7\textwidth}{!}{
    \begin{tabular}{
      >{\raggedright\arraybackslash}m{2.2cm}
      >{\centering\arraybackslash}m{1.8cm}
      >{\centering\arraybackslash}m{1.3cm}
      >{\centering\arraybackslash\bfseries}m{1.3cm}
      >{\centering\arraybackslash}m{1.8cm}
      >{\centering\arraybackslash}m{1.3cm}
      >{\centering\arraybackslash\bfseries}m{1.3cm}
    }
    \toprule
    \textbf{Model} & \multicolumn{3}{c}{\textbf{R² ($\uparrow$ higher is better)}} & \multicolumn{3}{c}{\textbf{MSE ($\times 10^{-2}$, $\downarrow$ lower is better)}} \\
    \cmidrule(lr){2-4} \cmidrule(lr){5-7}
    & \textbf{CO-OCCUR} & \textbf{MI} & \textbf{SMI} & \textbf{CO-OCCUR} & \textbf{MI} & \textbf{SMI} \\
    \midrule
    \multicolumn{7}{l}{\textbf{BLOOM}} \\
    bloom-560m & 0.384 & 0.589 & 0.591 & 8.566 & 7.729 & 7.723 \\
    bloom-1b1 & 0.577 & 0.715 & 0.719 & 9.123 & 7.950 & 7.939 \\
    bloom-1b7 & 0.584 & 0.740 & 0.743 & 9.389 & 8.083 & 8.072 \\
    bloom-3b & 0.705 & 0.782 & 0.786 & 9.314 & 7.863 & 7.849 \\
    bloom-7b1 & 0.676 & 0.804 & 0.807 & 9.943 & 8.309 & 8.296 \\
    bloom & 0.644 & 0.764 & 0.766 & 7.327 & 6.029 & 6.026 \\
    \midrule
    \multicolumn{7}{l}{\textbf{GPT-Neo}} \\
    gpt-neo-125m & 0.349 & 0.531 & 0.535 & 7.004 & 7.107 & 7.100 \\
    gpt-neo-1.3B & 0.671 & 0.810 & 0.814 & 8.233 & 7.444 & 7.427 \\
    gpt-neo-2.7B & 0.672 & 0.826 & 0.829 & 8.168 & 7.320 & 7.307 \\
    gpt-j-6b & 0.631 & 0.862 & 0.865 & 9.026 & 7.720 & 7.710 \\
    gpt-neox-20b & 0.593 & 0.841 & 0.843 & 8.557 & 7.476 & 7.469 \\
    \midrule
    \multicolumn{7}{l}{\textbf{Pythia}} \\
    pythia-14m & 0.240 & 0.375 & 0.378 & 6.855 & 7.183 & 7.177 \\
    pythia-31m & 0.287 & 0.381 & 0.384 & 6.905 & 7.202 & 7.196 \\
    pythia-70m & 0.355 & 0.430 & 0.434 & 6.749 & 7.051 & 7.044 \\
    pythia-160m & 0.454 & 0.522 & 0.526 & 6.877 & 6.969 & 6.961 \\
    pythia-410m & 0.524 & 0.701 & 0.705 & 7.439 & 7.035 & 7.024 \\
    pythia-1.4b & 0.574 & 0.799 & 0.803 & 8.254 & 7.325 & 7.312 \\
    pythia-2.8b & 0.631 & 0.845 & 0.848 & 8.593 & 7.505 & 7.491 \\
    pythia-6.9b & 0.637 & 0.855 & 0.859 & 8.538 & 7.373 & 7.360 \\
    pythia-12b & 0.602 & 0.853 & 0.856 & 8.741 & 7.514 & 7.503 \\
    \midrule
    \multicolumn{7}{l}{\textbf{TinyLlama}} \\
    TinyLlama-1.1B & 0.641 & 0.762 & 0.767 & 7.802 & 7.081 & 7.065 \\
    \midrule
    \multicolumn{7}{l}{\textbf{Ours}} \\
    ours-1.6b & 0.710 & 0.778 & 0.782 & 8.563 & 7.712 & 7.694 \\
    ours-7b & 0.784 & 0.837 & 0.839 & 8.982 & 8.028 & 8.016 \\
    ours-13b & 0.661 & 0.807 & 0.810 & 8.584 & 7.674 & 7.665 \\
    \bottomrule
    \end{tabular}
    }
    \end{center}
\end{table*}

\begin{figure*}[t]
\begin{center}
    \begin{minipage}{0.24\textwidth}
        \centering
        \includegraphics[width=\textwidth]{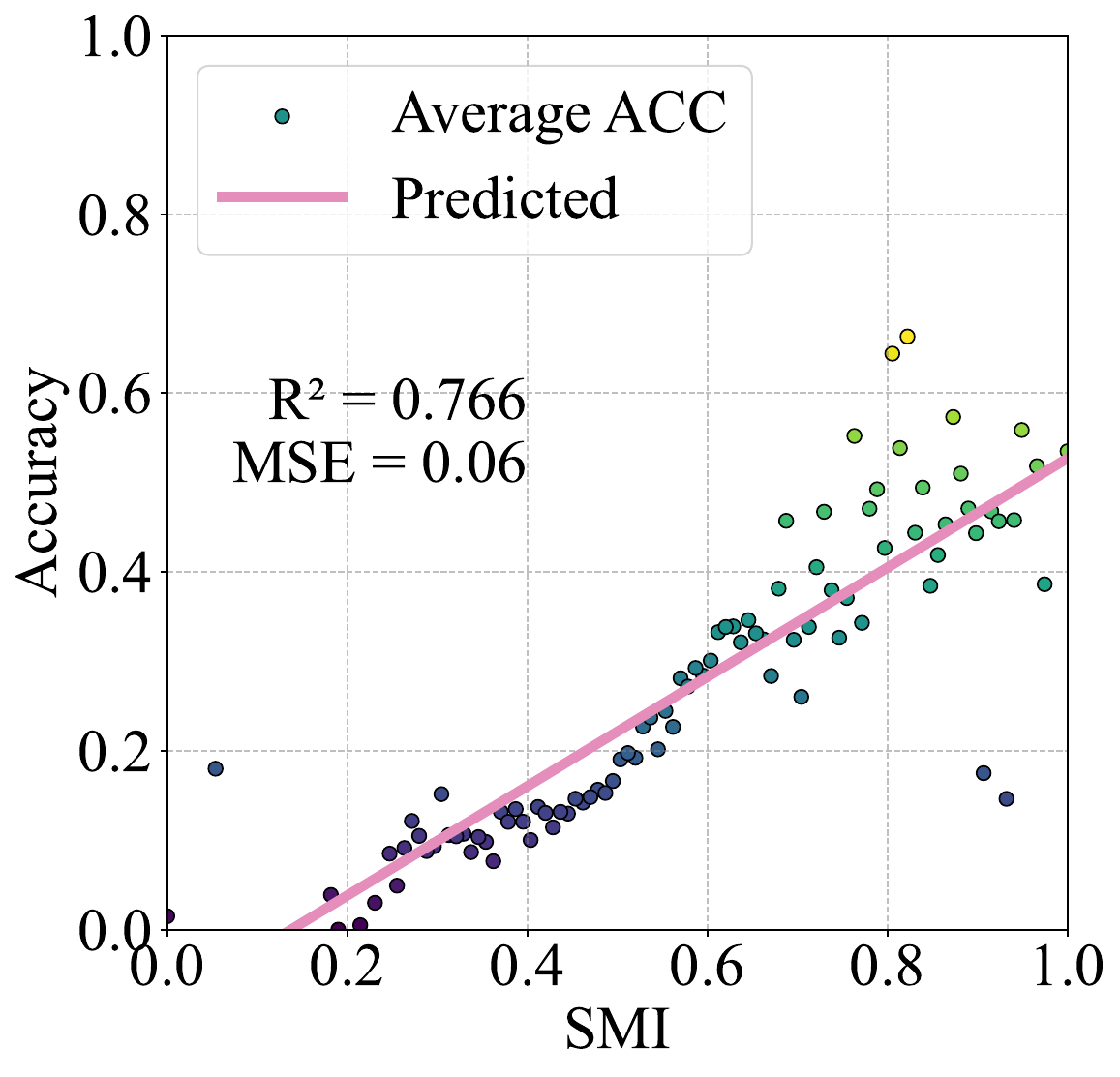} \\
        (a) bloom (176b)
    \end{minipage}
    \begin{minipage}{0.24\textwidth}
        \centering
        \includegraphics[width=\textwidth]{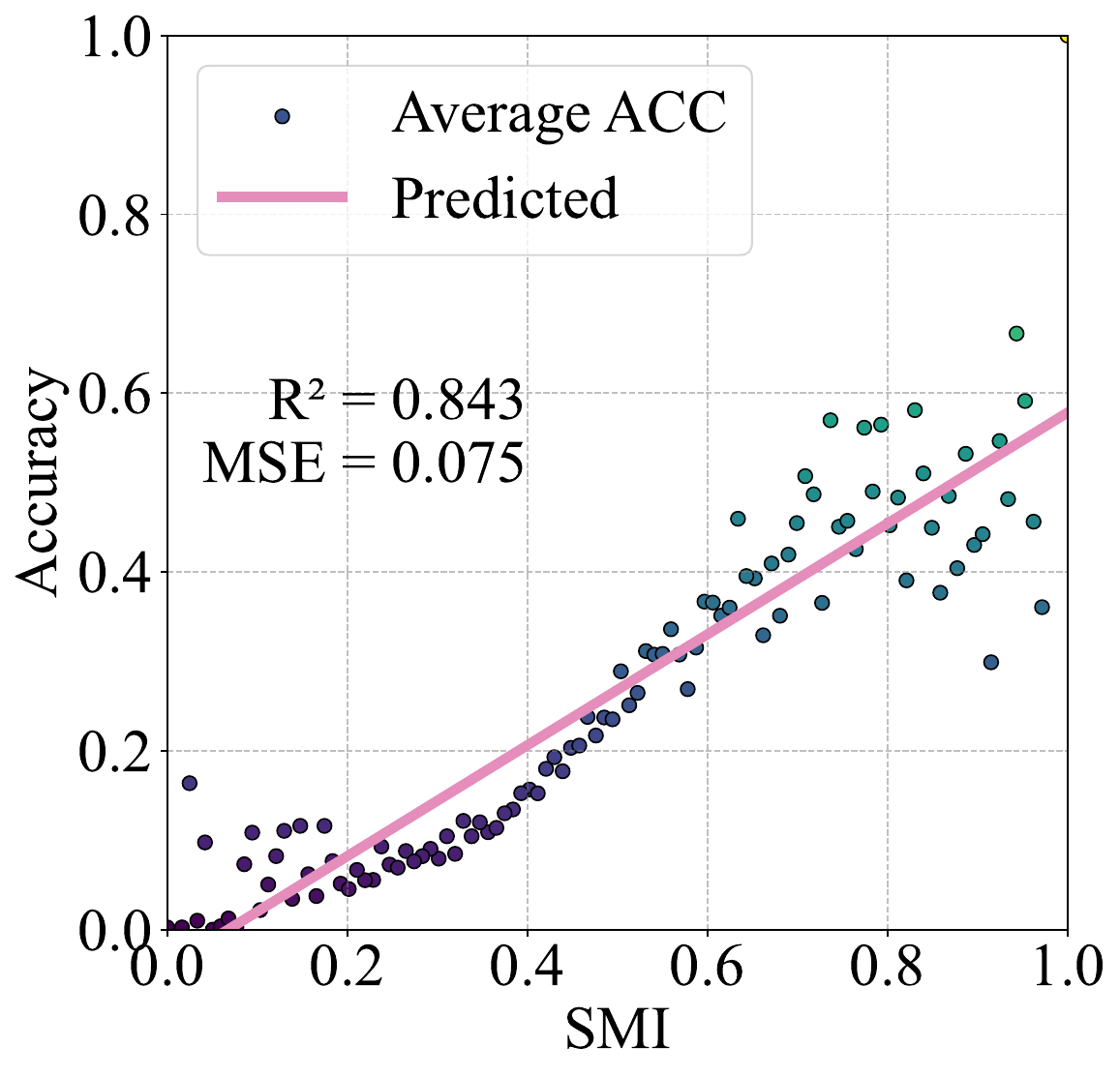} \\
        (b) gpt-neox-20b
    \end{minipage}
    \begin{minipage}{0.24\textwidth}
        \centering
        \includegraphics[width=\textwidth]{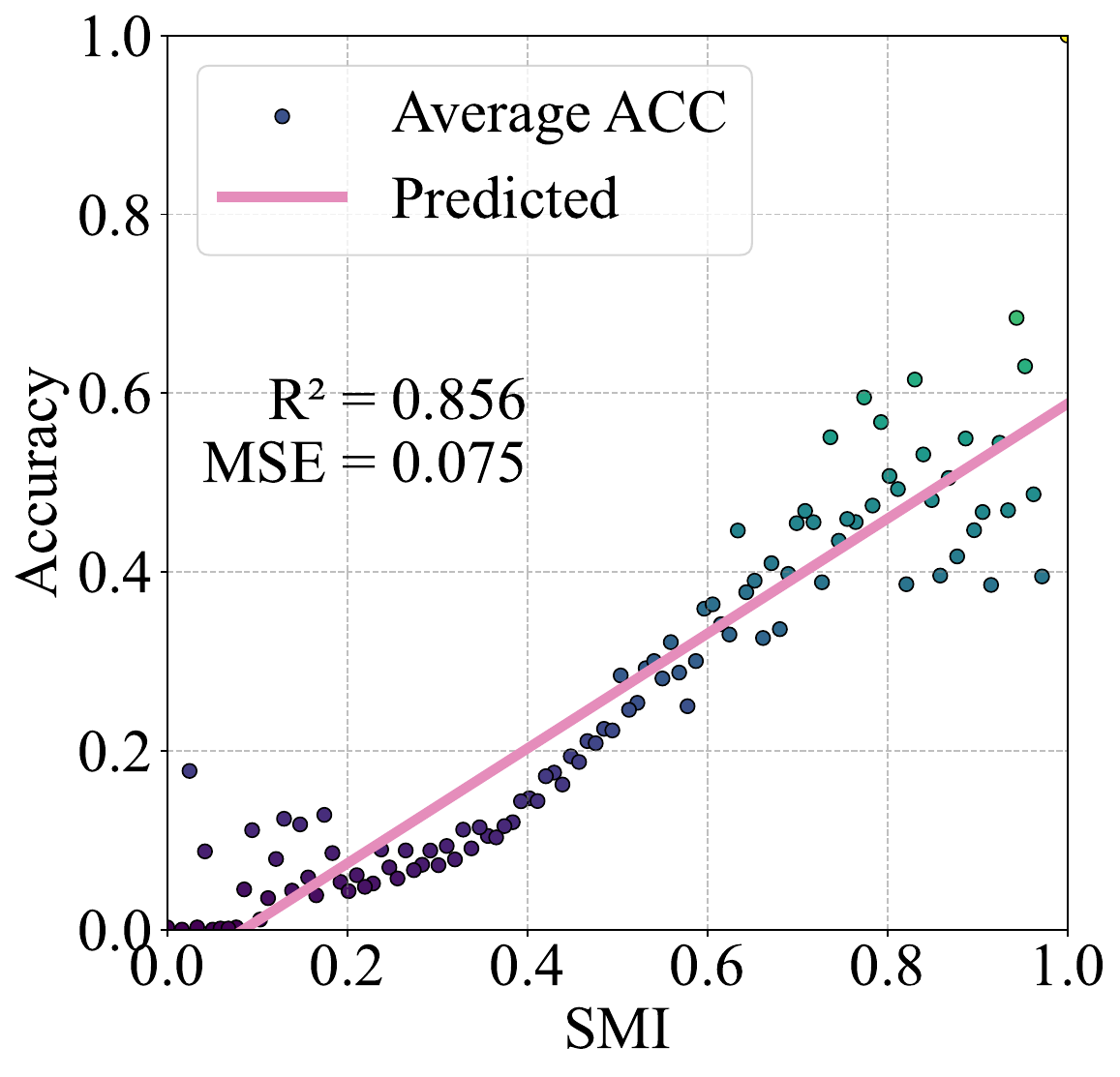} \\
        (c) pythia-12b
    \end{minipage}
    \begin{minipage}{0.24\textwidth}
        \centering
        \includegraphics[width=\textwidth]{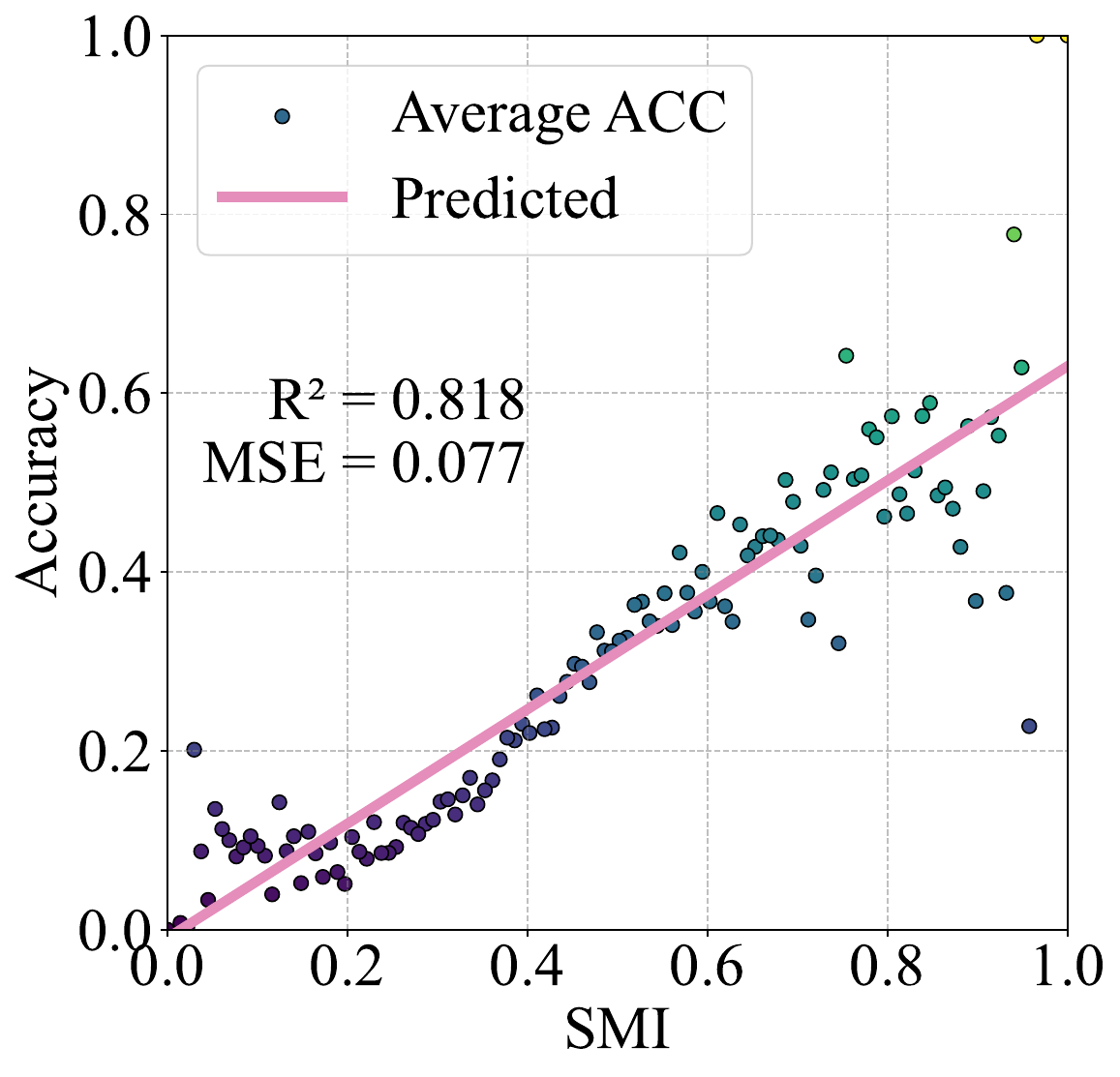} \\
        (d) ours-13b
    \end{minipage}
\caption{Relationship between the SMI metric and accuracy on QA tasks for different model series: (a) BLOOM (176B), (b) GPT-Neo (20B), (c) Pythia (12B), and (d) Ours (13B).}
\label{smi_figures}
\end{center}
\end{figure*}

\begin{figure}[t]
  \centering
  \includegraphics[width=0.6\columnwidth]{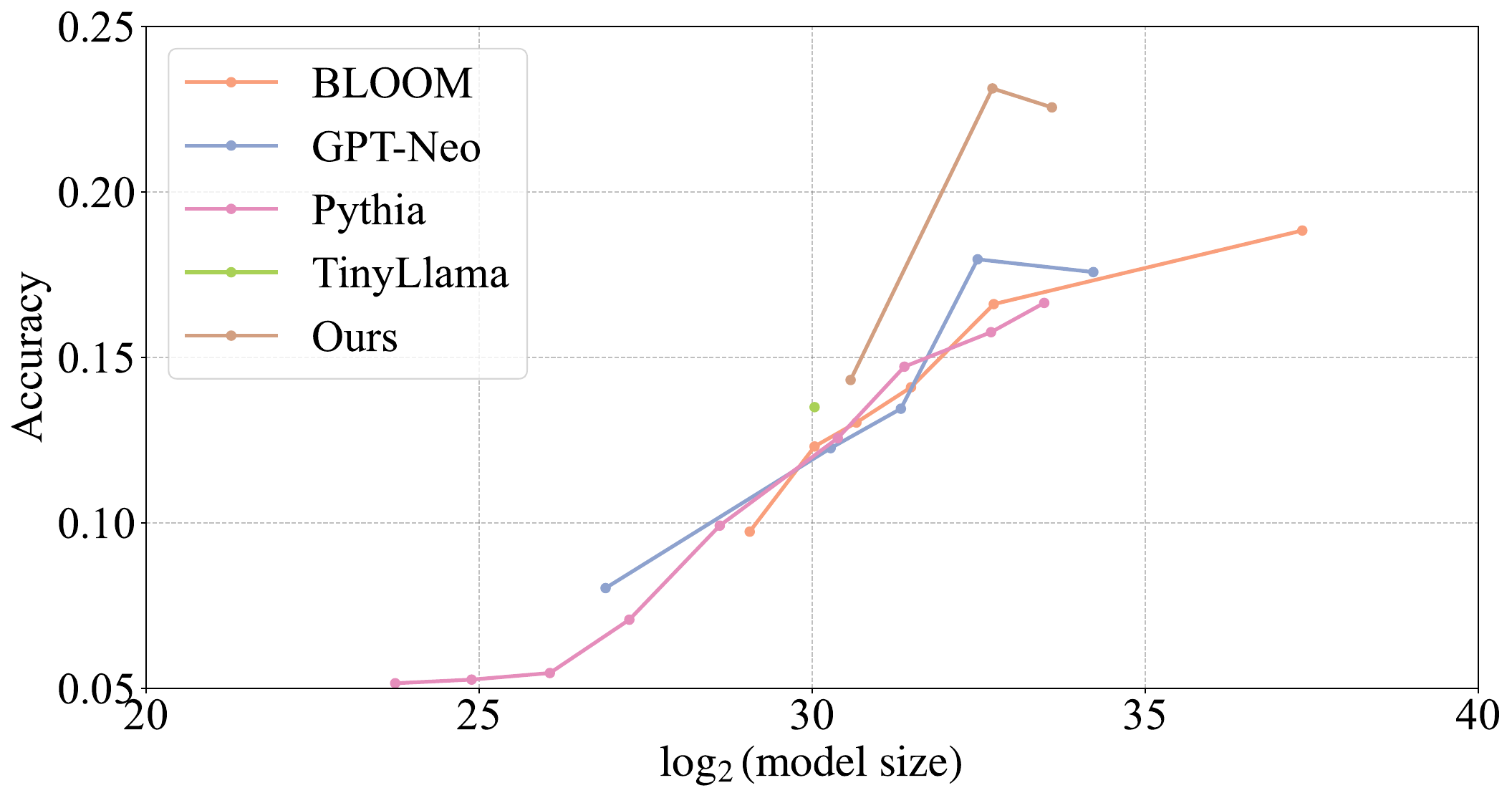}
  \caption{Relationship between average QA accuracy on the full evaluation dataset and model size across five different model families.}
  \label{pythia_acc}
\end{figure}

\subsection{Experimental Results}

\paragraph{Comparison of SMI, MI, and CO-OCCUR} 
Table~\ref{main_table} presents the R² and MSE results of CO-OCCUR, MI, and SMI across 24 models, with sizes ranging from 14M to 176B parameters. Overall, MI consistently improves over the co-occurrence baseline. For example, on \texttt{gpt-neo-2.7B}, R² increases from 0.672 (CO-OCCUR) to 0.829 (MI), while MSE decreases from 8.168 to 7.307. This suggests that jointly modeling frequency and specificity provides more informative signals for predicting closed-book QA performance. Incorporating model size into the metric (SMI) yields further gains across all models, though the magnitude of improvement is typically moderate. For instance, on \texttt{pythia-12b}, R² rises from 0.853 (MI) to 0.856 (SMI), and on \texttt{bloom}, from 0.764 to 0.766. These results suggest that incorporating model size into the metric provides a more principled account of the variance in observed QA accuracy.
Additional analysis and validation of SMI are detailed in Appendix~\ref{appendix:Additional Experimental Results}.

\paragraph{Diminishing returns from data scaling} 
Despite the improved predictive power of SMI, we find that knowledge retention is limited by the nature of pre-training data. As shown in Figure~\ref{smi_figures}, even when SMI reaches its maximum value of $1$, the closed-book QA accuracy does not exceed moderate levels. Specifically, the empirical upper bounds of accuracy for \texttt{bloom}, \texttt{gpt-neox-20b}, \texttt{pythia-12b}, and \texttt{ours-13b} are approximately $0.527$, $0.578$, $0.588$, and $0.630$, respectively. These results indicate that simply scaling the data cannot indefinitely improve factual recall.
Additional analyses are provided in the Appendix~\ref{appendix:Analysis in High-SMI Regions}.

\paragraph{Diminishing returns from model scaling} 
We further analyze the impact of model size on knowledge retention. Figure~\ref{pythia_acc} shows that the average accuracy on the full evaluation set scales approximately linearly with the logarithm of model parameters, which is consistent with prior scaling-law observations. However, the growth rate becomes increasingly shallow for very large models: even at $176$B parameters, the accuracy remains below $0.2$. This finding suggests that model scaling alone is insufficient for achieving high factual coverage and further motivates the exploration of complementary approaches, such as retrieval-augmented generation or external memory modules.
\section{Conclusion}
\label{conclusion}

In this work, we introduce a framework for quantifying knowledge retention in pre-trained language models via SMI, an information-theoretic predictor that integrates knowledge frequency, specificity, and model size. Our large-scale evaluation across 24 models shows that SMI achieves high predictive correlation ($R^2 > 0.7$ for models above 1B parameters) and remains robust across model families and domains. 

Beyond prediction, our analysis reveals diminishing returns from scaling data and model size, suggesting an inherent upper bound on parametric factual storage during pre-training. Furthermore, SMI serves as a diagnostic tool, indicating that while increasing co-occurrence frequency, refining data specificity, and scaling model size can enhance retention, achieving higher factual recall may necessitate complementary strategies such as retrieval-augmented generation, external memory modules, or targeted fine-tuning.

We hope this work provides a foundation for optimizing corpus composition and model design toward more effective factual knowledge retention.

\section{Limitations}
\label{limitations}

While our study provides an information-theoretic framework for predicting knowledge retention, several limitations remain.

First, our analysis is restricted to closed-book QA tasks over short factual triplets, as instantiated by the ParaRel dataset~\citep{DBLP:journals/tacl/ElazarKRRHSG21}. Accordingly, our notion of knowledge retention focuses on atomic, declarative facts and does not extend to multi-hop reasoning, non-atomic or unstructured knowledge, or procedural knowledge, all of which may involve substantially different memorization and retrieval dynamics.

Second, we rely on publicly available corpora and approximate document matching, which may introduce noise when estimating fact frequency and specificity.

Third, our experiments focus on models with up to tens of billions of parameters, leaving open the question of whether the observed trends hold for frontier-scale models with hundreds of billions of parameters.

Finally, our approach abstracts away training dynamics such as optimization schedules or curriculum learning, whose interactions with knowledge retention remain an important direction for future work.

\bibliography{main}

@inproceedings{DBLP:conf/nips/ChangPYYSCS24,
  author       = {Hoyeon Chang and
                  Jinho Park and
                  Seonghyeon Ye and
                  Sohee Yang and
                  Youngkyung Seo and
                  Du{-}Seong Chang and
                  Minjoon Seo},
  editor       = {Amir Globersons and
                  Lester Mackey and
                  Danielle Belgrave and
                  Angela Fan and
                  Ulrich Paquet and
                  Jakub M. Tomczak and
                  Cheng Zhang},
  title        = {How Do Large Language Models Acquire Factual Knowledge During Pretraining?},
  booktitle    = {Advances in Neural Information Processing Systems 38: Annual Conference
                  on Neural Information Processing Systems 2024, NeurIPS 2024, Vancouver,
                  BC, Canada, December 10 - 15, 2024},
  year         = {2024},
  url          = {http://papers.nips.cc/paper\_files/paper/2024/hash/6fdf57c71bc1f1ee29014b8dc52e723f-Abstract-Conference.html},
  bibsource    = {dblp computer science bibliography, https://dblp.org}
}

@article{zhang2025llmeval,
  title={LLMEval-3: A Large-Scale Longitudinal Study on Robust and Fair Evaluation of Large Language Models},
  author={Zhang, Ming and Shen, Yujiong and Deng, Jingyi and Wang, Yuhui and Zhang, Yue and Wang, Junzhe and Liu, Shichun and Dou, Shihan and Sha, Huayu and Peng, Qiyuan and others},
  journal={arXiv preprint arXiv:2508.05452},
  year={2025}
}

@article{alabdulmohsin2022revisiting,
  title={Revisiting neural scaling laws in language and vision},
  author={Alabdulmohsin, Ibrahim M and Neyshabur, Behnam and Zhai, Xiaohua},
  journal={Advances in Neural Information Processing Systems},
  volume={35},
  pages={22300--22312},
  year={2022}
}

@article{ruan2405observational,
  title={Observational scaling laws and the predictability of language model performance, 2024},
  author={Ruan, Yangjun and Maddison, Chris J and Hashimoto, Tatsunori},
  journal={URL https://arxiv. org/abs/2405.10938}
}

@article{biderman2023emergent,
  title={Emergent and predictable memorization in large language models},
  author={Biderman, Stella and Prashanth, Usvsn and Sutawika, Lintang and Schoelkopf, Hailey and Anthony, Quentin and Purohit, Shivanshu and Raff, Edward},
  journal={Advances in Neural Information Processing Systems},
  volume={36},
  pages={28072--28090},
  year={2023}
}

@inproceedings{carlini2021extracting,
  title={Extracting training data from large language models},
  author={Carlini, Nicholas and Tramer, Florian and Wallace, Eric and Jagielski, Matthew and Herbert-Voss, Ariel and Lee, Katherine and Roberts, Adam and Brown, Tom and Song, Dawn and Erlingsson, Ulfar and others},
  booktitle={30th USENIX security symposium (USENIX Security 21)},
  pages={2633--2650},
  year={2021}
}

@article{hoffmann2022training,
  title={Training compute-optimal large language models},
  author={Hoffmann, Jordan and Borgeaud, Sebastian and Mensch, Arthur and Buchatskaya, Elena and Cai, Trevor and Rutherford, Eliza and Casas, Diego de Las and Hendricks, Lisa Anne and Welbl, Johannes and Clark, Aidan and others},
  journal={arXiv preprint arXiv:2203.15556},
  year={2022}
}

@inproceedings{DBLP:conf/nips/VaswaniSPUJGKP17,
  author       = {Ashish Vaswani and
                  Noam Shazeer and
                  Niki Parmar and
                  Jakob Uszkoreit and
                  Llion Jones and
                  Aidan N. Gomez and
                  Lukasz Kaiser and
                  Illia Polosukhin},
  editor       = {Isabelle Guyon and
                  Ulrike von Luxburg and
                  Samy Bengio and
                  Hanna M. Wallach and
                  Rob Fergus and
                  S. V. N. Vishwanathan and
                  Roman Garnett},
  title        = {Attention is All you Need},
  booktitle    = {Advances in Neural Information Processing Systems 30: Annual Conference
                  on Neural Information Processing Systems 2017, December 4-9, 2017,
                  Long Beach, CA, {USA}},
  pages        = {5998--6008},
  year         = {2017},
  url          = {https://proceedings.neurips.cc/paper/2017/hash/3f5ee243547dee91fbd053c1c4a845aa-Abstract.html},
  bibsource    = {dblp computer science bibliography, https://dblp.org}
}

@inproceedings{DBLP:conf/emnlp/RobertsRS20,
  author       = {Adam Roberts and
                  Colin Raffel and
                  Noam Shazeer},
  editor       = {Bonnie Webber and
                  Trevor Cohn and
                  Yulan He and
                  Yang Liu},
  title        = {How Much Knowledge Can You Pack Into the Parameters of a Language
                  Model?},
  booktitle    = {Proceedings of the 2020 Conference on Empirical Methods in Natural
                  Language Processing, {EMNLP} 2020, Online, November 16-20, 2020},
  pages        = {5418--5426},
  publisher    = {Association for Computational Linguistics},
  year         = {2020},
  url          = {https://doi.org/10.18653/v1/2020.emnlp-main.437},
  doi          = {10.18653/V1/2020.EMNLP-MAIN.437},
  bibsource    = {dblp computer science bibliography, https://dblp.org}
}

@article{DBLP:journals/corr/abs-2211-05100,
  author       = {Teven Le Scao and
                  Angela Fan and
                  Christopher Akiki and
                  Ellie Pavlick and
                  Suzana Ilic and
                  Daniel Hesslow and
                  Roman Castagn{\'{e}} and
                  Alexandra Sasha Luccioni and
                  Fran{\c{c}}ois Yvon and
                  Matthias Gall{\'{e}} and
                  Jonathan Tow and
                  Alexander M. Rush and
                  Stella Biderman and
                  Albert Webson and
                  Pawan Sasanka Ammanamanchi and
                  Thomas Wang and
                  Beno{\^{\i}}t Sagot and
                  Niklas Muennighoff and
                  Albert Villanova del Moral and
                  Olatunji Ruwase and
                  Rachel Bawden and
                  Stas Bekman and
                  Angelina McMillan{-}Major and
                  Iz Beltagy and
                  Huu Nguyen and
                  Lucile Saulnier and
                  Samson Tan and
                  Pedro Ortiz Suarez and
                  Victor Sanh and
                  Hugo Lauren{\c{c}}on and
                  Yacine Jernite and
                  Julien Launay and
                  Margaret Mitchell and
                  Colin Raffel and
                  Aaron Gokaslan and
                  Adi Simhi and
                  Aitor Soroa and
                  Alham Fikri Aji and
                  Amit Alfassy and
                  Anna Rogers and
                  Ariel Kreisberg Nitzav and
                  Canwen Xu and
                  Chenghao Mou and
                  Chris Emezue and
                  Christopher Klamm and
                  Colin Leong and
                  Daniel van Strien and
                  David Ifeoluwa Adelani and
                  et al.},
  title        = {{BLOOM:} {A} 176B-Parameter Open-Access Multilingual Language Model},
  journal      = {CoRR},
  volume       = {abs/2211.05100},
  year         = {2022},
  url          = {https://doi.org/10.48550/arXiv.2211.05100},
  doi          = {10.48550/ARXIV.2211.05100},
  eprinttype    = {arXiv},
  eprint       = {2211.05100},
  bibsource    = {dblp computer science bibliography, https://dblp.org}
}

@inproceedings{DBLP:conf/icml/BidermanSABOHKP23,
  author       = {Stella Biderman and
                  Hailey Schoelkopf and
                  Quentin Gregory Anthony and
                  Herbie Bradley and
                  Kyle O'Brien and
                  Eric Hallahan and
                  Mohammad Aflah Khan and
                  Shivanshu Purohit and
                  USVSN Sai Prashanth and
                  Edward Raff and
                  Aviya Skowron and
                  Lintang Sutawika and
                  Oskar van der Wal},
  editor       = {Andreas Krause and
                  Emma Brunskill and
                  Kyunghyun Cho and
                  Barbara Engelhardt and
                  Sivan Sabato and
                  Jonathan Scarlett},
  title        = {Pythia: {A} Suite for Analyzing Large Language Models Across Training
                  and Scaling},
  booktitle    = {International Conference on Machine Learning, {ICML} 2023, 23-29 July
                  2023, Honolulu, Hawaii, {USA}},
  series       = {Proceedings of Machine Learning Research},
  volume       = {202},
  pages        = {2397--2430},
  publisher    = {{PMLR}},
  year         = {2023},
  url          = {https://proceedings.mlr.press/v202/biderman23a.html},
  bibsource    = {dblp computer science bibliography, https://dblp.org}
}

@misc{mesh-transformer-jax,
  author = {Wang, Ben},
  title = {{Mesh-Transformer-JAX: Model-Parallel Implementation of Transformer Language Model with JAX}},
  howpublished = {\url{https://github.com/kingoflolz/mesh-transformer-jax}},
  year = 2021,
  month = May
}

@article{DBLP:journals/corr/abs-2204-06745,
  author       = {Sid Black and
                  Stella Biderman and
                  Eric Hallahan and
                  Quentin Anthony and
                  Leo Gao and
                  Laurence Golding and
                  Horace He and
                  Connor Leahy and
                  Kyle McDonell and
                  Jason Phang and
                  Michael Pieler and
                  USVSN Sai Prashanth and
                  Shivanshu Purohit and
                  Laria Reynolds and
                  Jonathan Tow and
                  Ben Wang and
                  Samuel Weinbach},
  title        = {GPT-NeoX-20B: An Open-Source Autoregressive Language Model},
  journal      = {CoRR},
  volume       = {abs/2204.06745},
  year         = {2022},
  url          = {https://doi.org/10.48550/arXiv.2204.06745},
  doi          = {10.48550/ARXIV.2204.06745},
  eprinttype    = {arXiv},
  eprint       = {2204.06745},
  bibsource    = {dblp computer science bibliography, https://dblp.org}
}

@software{gpt-neo,
  author       = {Black, Sid and
                  Leo, Gao and
                  Wang, Phil and
                  Leahy, Connor and
                  Biderman, Stella},
  title        = {{GPT-Neo: Large Scale Autoregressive Language 
                   Modeling with Mesh-Tensorflow}},
  month        = mar,
  year         = 2021,
  note         = {{If you use this software, please cite it using 
                   these metadata.}},
  publisher    = {Zenodo},
  version      = {1.0},
  doi          = {10.5281/zenodo.5297715},
  url          = {https://doi.org/10.5281/zenodo.5297715}
}

@ONLINE{wikidump,
    author = "Wikimedia Foundation",
    title  = "Wikimedia Downloads",
    url    = "https://dumps.wikimedia.org"
}

@inproceedings{DBLP:conf/nips/PenedoMHCACPAL23,
  author       = {Guilherme Penedo and
                  Quentin Malartic and
                  Daniel Hesslow and
                  Ruxandra Cojocaru and
                  Hamza Alobeidli and
                  Alessandro Cappelli and
                  Baptiste Pannier and
                  Ebtesam Almazrouei and
                  Julien Launay},
  editor       = {Alice Oh and
                  Tristan Naumann and
                  Amir Globerson and
                  Kate Saenko and
                  Moritz Hardt and
                  Sergey Levine},
  title        = {The RefinedWeb Dataset for Falcon {LLM:} Outperforming Curated Corpora
                  with Web Data Only},
  booktitle    = {Advances in Neural Information Processing Systems 36: Annual Conference
                  on Neural Information Processing Systems 2023, NeurIPS 2023, New Orleans,
                  LA, USA, December 10 - 16, 2023},
  year         = {2023},
  url          = {http://papers.nips.cc/paper\_files/paper/2023/hash/fa3ed726cc5073b9c31e3e49a807789c-Abstract-Datasets\_and\_Benchmarks.html},
  bibsource    = {dblp computer science bibliography, https://dblp.org}
}

@misc{cerebras2023slimpajama,
  author = {Soboleva, Daria and Al-Khateeb, Faisal and Myers, Robert and Steeves, Jacob R and Hestness, Joel and Dey, Nolan},
  title = {{SlimPajama: A 627B token cleaned and deduplicated version of RedPajama}},
  month = {June},
  year = 2023,
  howpublished = {https://www.cerebras.net/blog/slimpajama-a-627b-token-cleaned-and-deduplicated-version-of-redpajama},
  url = {https://huggingface.co/datasets/cerebras/SlimPajama-627B},
}

@inproceedings{DBLP:conf/nips/LaurenconSWAMSW22,
  author       = {Hugo Lauren{\c{c}}on and
                  Lucile Saulnier and
                  Thomas Wang and
                  Christopher Akiki and
                  Albert Villanova del Moral and
                  Teven Le Scao and
                  Leandro von Werra and
                  Chenghao Mou and
                  Eduardo Gonz{\'{a}}lez Ponferrada and
                  Huu Nguyen and
                  J{\"{o}}rg Frohberg and
                  Mario Sasko and
                  Quentin Lhoest and
                  Angelina McMillan{-}Major and
                  G{\'{e}}rard Dupont and
                  Stella Biderman and
                  Anna Rogers and
                  Loubna Ben Allal and
                  Francesco De Toni and
                  Giada Pistilli and
                  Olivier Nguyen and
                  Somaieh Nikpoor and
                  Maraim Masoud and
                  Pierre Colombo and
                  Javier de la Rosa and
                  Paulo Villegas and
                  Tristan Thrush and
                  Shayne Longpre and
                  Sebastian Nagel and
                  Leon Weber and
                  Manuel Mu{\~{n}}oz and
                  Jian Zhu and
                  Daniel van Strien and
                  Zaid Alyafeai and
                  Khalid Almubarak and
                  Minh Chien Vu and
                  Itziar Gonzalez{-}Dios and
                  Aitor Soroa and
                  Kyle Lo and
                  Manan Dey and
                  Pedro Ortiz Suarez and
                  Aaron Gokaslan and
                  Shamik Bose and
                  David Ifeoluwa Adelani and
                  Long Phan and
                  Hieu Tran and
                  Ian Yu and
                  Suhas Pai and
                  Jenny Chim and
                  Violette Lepercq and
                  Suzana Ilic and
                  Margaret Mitchell and
                  Alexandra Sasha Luccioni and
                  Yacine Jernite},
  editor       = {Sanmi Koyejo and
                  S. Mohamed and
                  A. Agarwal and
                  Danielle Belgrave and
                  K. Cho and
                  A. Oh},
  title        = {The BigScience {ROOTS} Corpus: {A} 1.6TB Composite Multilingual Dataset},
  booktitle    = {Advances in Neural Information Processing Systems 35: Annual Conference
                  on Neural Information Processing Systems 2022, NeurIPS 2022, New Orleans,
                  LA, USA, November 28 - December 9, 2022},
  year         = {2022},
  url          = {http://papers.nips.cc/paper\_files/paper/2022/hash/ce9e92e3de2372a4b93353eb7f3dc0bd-Abstract-Datasets\_and\_Benchmarks.html},
  bibsource    = {dblp computer science bibliography, https://dblp.org}
}

@article{DBLP:journals/corr/abs-2101-00027,
  author       = {Leo Gao and
                  Stella Biderman and
                  Sid Black and
                  Laurence Golding and
                  Travis Hoppe and
                  Charles Foster and
                  Jason Phang and
                  Horace He and
                  Anish Thite and
                  Noa Nabeshima and
                  Shawn Presser and
                  Connor Leahy},
  title        = {The Pile: An 800GB Dataset of Diverse Text for Language Modeling},
  journal      = {CoRR},
  volume       = {abs/2101.00027},
  year         = {2021},
  url          = {https://arxiv.org/abs/2101.00027},
  eprinttype    = {arXiv},
  eprint       = {2101.00027},
  bibsource    = {dblp computer science bibliography, https://dblp.org}
}

@inproceedings{DBLP:conf/icml/KandpalDRWR23,
  author       = {Nikhil Kandpal and
                  Haikang Deng and
                  Adam Roberts and
                  Eric Wallace and
                  Colin Raffel},
  editor       = {Andreas Krause and
                  Emma Brunskill and
                  Kyunghyun Cho and
                  Barbara Engelhardt and
                  Sivan Sabato and
                  Jonathan Scarlett},
  title        = {Large Language Models Struggle to Learn Long-Tail Knowledge},
  booktitle    = {International Conference on Machine Learning, {ICML} 2023, 23-29 July
                  2023, Honolulu, Hawaii, {USA}},
  series       = {Proceedings of Machine Learning Research},
  volume       = {202},
  pages        = {15696--15707},
  publisher    = {{PMLR}},
  year         = {2023},
  url          = {https://proceedings.mlr.press/v202/kandpal23a.html},
  bibsource    = {dblp computer science bibliography, https://dblp.org}
}

@inproceedings{DBLP:conf/acl/MallenAZDKH23,
  author       = {Alex Mallen and
                  Akari Asai and
                  Victor Zhong and
                  Rajarshi Das and
                  Daniel Khashabi and
                  Hannaneh Hajishirzi},
  editor       = {Anna Rogers and
                  Jordan L. Boyd{-}Graber and
                  Naoaki Okazaki},
  title        = {When Not to Trust Language Models: Investigating Effectiveness of
                  Parametric and Non-Parametric Memories},
  booktitle    = {Proceedings of the 61st Annual Meeting of the Association for Computational
                  Linguistics (Volume 1: Long Papers), {ACL} 2023, Toronto, Canada,
                  July 9-14, 2023},
  pages        = {9802--9822},
  publisher    = {Association for Computational Linguistics},
  year         = {2023},
  url          = {https://doi.org/10.18653/v1/2023.acl-long.546},
  doi          = {10.18653/V1/2023.ACL-LONG.546},
  bibsource    = {dblp computer science bibliography, https://dblp.org}
}

@article{DBLP:journals/tmlr/LiAZMKMMALCLZZW23,
  author       = {Raymond Li and
                  Loubna Ben Allal and
                  Yangtian Zi and
                  Niklas Muennighoff and
                  Denis Kocetkov and
                  Chenghao Mou and
                  Marc Marone and
                  Christopher Akiki and
                  Jia Li and
                  Jenny Chim and
                  Qian Liu and
                  Evgenii Zheltonozhskii and
                  Terry Yue Zhuo and
                  Thomas Wang and
                  Olivier Dehaene and
                  Mishig Davaadorj and
                  Joel Lamy{-}Poirier and
                  Jo{\~{a}}o Monteiro and
                  Oleh Shliazhko and
                  Nicolas Gontier and
                  Nicholas Meade and
                  Armel Zebaze and
                  Ming{-}Ho Yee and
                  Logesh Kumar Umapathi and
                  Jian Zhu and
                  Benjamin Lipkin and
                  Muhtasham Oblokulov and
                  Zhiruo Wang and
                  Rudra Murthy V and
                  Jason T. Stillerman and
                  Siva Sankalp Patel and
                  Dmitry Abulkhanov and
                  Marco Zocca and
                  Manan Dey and
                  Zhihan Zhang and
                  Nour Fahmy and
                  Urvashi Bhattacharyya and
                  Wenhao Yu and
                  Swayam Singh and
                  Sasha Luccioni and
                  Paulo Villegas and
                  Maxim Kunakov and
                  Fedor Zhdanov and
                  Manuel Romero and
                  Tony Lee and
                  Nadav Timor and
                  Jennifer Ding and
                  Claire Schlesinger and
                  Hailey Schoelkopf and
                  Jan Ebert and
                  Tri Dao and
                  Mayank Mishra and
                  Alex Gu and
                  Jennifer Robinson and
                  Carolyn Jane Anderson and
                  Brendan Dolan{-}Gavitt and
                  Danish Contractor and
                  Siva Reddy and
                  Daniel Fried and
                  Dzmitry Bahdanau and
                  Yacine Jernite and
                  Carlos Mu{\~{n}}oz Ferrandis and
                  Sean Hughes and
                  Thomas Wolf and
                  Arjun Guha and
                  Leandro von Werra and
                  Harm de Vries},
  title        = {StarCoder: may the source be with you!},
  journal      = {Trans. Mach. Learn. Res.},
  volume       = {2023},
  year         = {2023},
  url          = {https://openreview.net/forum?id=KoFOg41haE},
  bibsource    = {dblp computer science bibliography, https://dblp.org}
}

@inproceedings{DBLP:conf/emnlp/PetroniRRLBWM19,
  author       = {Fabio Petroni and
                  Tim Rockt{\"{a}}schel and
                  Sebastian Riedel and
                  Patrick S. H. Lewis and
                  Anton Bakhtin and
                  Yuxiang Wu and
                  Alexander H. Miller},
  editor       = {Kentaro Inui and
                  Jing Jiang and
                  Vincent Ng and
                  Xiaojun Wan},
  title        = {Language Models as Knowledge Bases?},
  booktitle    = {Proceedings of the 2019 Conference on Empirical Methods in Natural
                  Language Processing and the 9th International Joint Conference on
                  Natural Language Processing, {EMNLP-IJCNLP} 2019, Hong Kong, China,
                  November 3-7, 2019},
  pages        = {2463--2473},
  publisher    = {Association for Computational Linguistics},
  year         = {2019},
  url          = {https://doi.org/10.18653/v1/D19-1250},
  doi          = {10.18653/V1/D19-1250},
  bibsource    = {dblp computer science bibliography, https://dblp.org}
}

@article{DBLP:journals/corr/abs-2303-08774,
  author       = {OpenAI},
  title        = {{GPT-4} Technical Report},
  journal      = {CoRR},
  volume       = {abs/2303.08774},
  year         = {2023},
  url          = {https://doi.org/10.48550/arXiv.2303.08774},
  doi          = {10.48550/ARXIV.2303.08774},
  eprinttype    = {arXiv},
  eprint       = {2303.08774},
  bibsource    = {dblp computer science bibliography, https://dblp.org}
}

@article{DBLP:journals/corr/abs-2401-02385,
  author       = {Peiyuan Zhang and
                  Guangtao Zeng and
                  Tianduo Wang and
                  Wei Lu},
  title        = {TinyLlama: An Open-Source Small Language Model},
  journal      = {CoRR},
  volume       = {abs/2401.02385},
  year         = {2024},
  url          = {https://doi.org/10.48550/arXiv.2401.02385},
  doi          = {10.48550/ARXIV.2401.02385},
  eprinttype    = {arXiv},
  eprint       = {2401.02385},
  bibsource    = {dblp computer science bibliography, https://dblp.org}
}

@inproceedings{DBLP:conf/icml/Allen-ZhuL24,
  author       = {Zeyuan Allen{-}Zhu and
                  Yuanzhi Li},
  title        = {Physics of Language Models: Part 3.1, Knowledge Storage and Extraction},
  booktitle    = {Forty-first International Conference on Machine Learning, {ICML} 2024,
                  Vienna, Austria, July 21-27, 2024},
  publisher    = {OpenReview.net},
  year         = {2024},
  url          = {https://openreview.net/forum?id=5x788rqbcj},
  bibsource    = {dblp computer science bibliography, https://dblp.org}
}

@article{DBLP:journals/jmlr/ChowdheryNDBMRBCSGSSTMRBTSPRDHPBAI23,
  author       = {Aakanksha Chowdhery and
                  Sharan Narang and
                  Jacob Devlin and
                  Maarten Bosma and
                  Gaurav Mishra and
                  Adam Roberts and
                  Paul Barham and
                  Hyung Won Chung and
                  Charles Sutton and
                  Sebastian Gehrmann and
                  Parker Schuh and
                  Kensen Shi and
                  Sasha Tsvyashchenko and
                  Joshua Maynez and
                  Abhishek Rao and
                  Parker Barnes and
                  Yi Tay and
                  Noam Shazeer and
                  Vinodkumar Prabhakaran and
                  Emily Reif and
                  Nan Du and
                  Ben Hutchinson and
                  Reiner Pope and
                  James Bradbury and
                  Jacob Austin and
                  Michael Isard and
                  Guy Gur{-}Ari and
                  Pengcheng Yin and
                  Toju Duke and
                  Anselm Levskaya and
                  Sanjay Ghemawat and
                  Sunipa Dev and
                  Henryk Michalewski and
                  Xavier Garcia and
                  Vedant Misra and
                  Kevin Robinson and
                  Liam Fedus and
                  Denny Zhou and
                  Daphne Ippolito and
                  David Luan and
                  Hyeontaek Lim and
                  Barret Zoph and
                  Alexander Spiridonov and
                  Ryan Sepassi and
                  David Dohan and
                  Shivani Agrawal and
                  Mark Omernick and
                  Andrew M. Dai and
                  Thanumalayan Sankaranarayana Pillai and
                  Marie Pellat and
                  Aitor Lewkowycz and
                  Erica Moreira and
                  Rewon Child and
                  Oleksandr Polozov and
                  Katherine Lee and
                  Zongwei Zhou and
                  Xuezhi Wang and
                  Brennan Saeta and
                  Mark Diaz and
                  Orhan Firat and
                  Michele Catasta and
                  Jason Wei and
                  Kathy Meier{-}Hellstern and
                  Douglas Eck and
                  Jeff Dean and
                  Slav Petrov and
                  Noah Fiedel},
  title        = {PaLM: Scaling Language Modeling with Pathways},
  journal      = {J. Mach. Learn. Res.},
  volume       = {24},
  pages        = {240:1--240:113},
  year         = {2023},
  url          = {https://jmlr.org/papers/v24/22-1144.html},
  bibsource    = {dblp computer science bibliography, https://dblp.org}
}

@article{DBLP:journals/corr/abs-2311-00871,
  author       = {Steve Yadlowsky and
                  Lyric Doshi and
                  Nilesh Tripuraneni},
  title        = {Pretraining Data Mixtures Enable Narrow Model Selection Capabilities
                  in Transformer Models},
  journal      = {CoRR},
  volume       = {abs/2311.00871},
  year         = {2023},
  url          = {https://doi.org/10.48550/arXiv.2311.00871},
  doi          = {10.48550/ARXIV.2311.00871},
  eprinttype    = {arXiv},
  eprint       = {2311.00871},
  bibsource    = {dblp computer science bibliography, https://dblp.org}
}

@article{DBLP:journals/corr/abs-2309-13638,
  author       = {R. Thomas McCoy and
                  Shunyu Yao and
                  Dan Friedman and
                  Matthew Hardy and
                  Thomas L. Griffiths},
  title        = {Embers of Autoregression: Understanding Large Language Models Through
                  the Problem They are Trained to Solve},
  journal      = {CoRR},
  volume       = {abs/2309.13638},
  year         = {2023},
  url          = {https://doi.org/10.48550/arXiv.2309.13638},
  doi          = {10.48550/ARXIV.2309.13638},
  eprinttype    = {arXiv},
  eprint       = {2309.13638},
  bibsource    = {dblp computer science bibliography, https://dblp.org}
}

@inproceedings{DBLP:conf/emnlp/RazeghiL0022,
  author       = {Yasaman Razeghi and
                  Robert L. Logan IV and
                  Matt Gardner and
                  Sameer Singh},
  editor       = {Yoav Goldberg and
                  Zornitsa Kozareva and
                  Yue Zhang},
  title        = {Impact of Pretraining Term Frequencies on Few-Shot Numerical Reasoning},
  booktitle    = {Findings of the Association for Computational Linguistics: {EMNLP}
                  2022, Abu Dhabi, United Arab Emirates, December 7-11, 2022},
  pages        = {840--854},
  publisher    = {Association for Computational Linguistics},
  year         = {2022},
  url          = {https://doi.org/10.18653/v1/2022.findings-emnlp.59},
  doi          = {10.18653/V1/2022.FINDINGS-EMNLP.59},
  bibsource    = {dblp computer science bibliography, https://dblp.org}
}

@inproceedings{DBLP:conf/iclr/CarliniIJLTZ23,
  author       = {Nicholas Carlini and
                  Daphne Ippolito and
                  Matthew Jagielski and
                  Katherine Lee and
                  Florian Tram{\`{e}}r and
                  Chiyuan Zhang},
  title        = {Quantifying Memorization Across Neural Language Models},
  booktitle    = {The Eleventh International Conference on Learning Representations,
                  {ICLR} 2023, Kigali, Rwanda, May 1-5, 2023},
  publisher    = {OpenReview.net},
  year         = {2023},
  url          = {https://openreview.net/forum?id=TatRHT\_1cK},
  bibsource    = {dblp computer science bibliography, https://dblp.org}
}

@article{DBLP:journals/corr/abs-2402-14273,
  author       = {Qiyuan He and
                  Yizhong Wang and
                  Wenya Wang},
  title        = {Can Language Models Act as Knowledge Bases at Scale?},
  journal      = {CoRR},
  volume       = {abs/2402.14273},
  year         = {2024},
  url          = {https://doi.org/10.48550/arXiv.2402.14273},
  doi          = {10.48550/ARXIV.2402.14273},
  eprinttype    = {arXiv},
  eprint       = {2402.14273},
  bibsource    = {dblp computer science bibliography, https://dblp.org}
}

@inproceedings{DBLP:conf/acl/JuCY0DZL24,
  author       = {Tianjie Ju and
                  Yijin Chen and
                  Xinwei Yuan and
                  Zhuosheng Zhang and
                  Wei Du and
                  Yubin Zheng and
                  Gongshen Liu},
  editor       = {Lun{-}Wei Ku and
                  Andre Martins and
                  Vivek Srikumar},
  title        = {Investigating Multi-Hop Factual Shortcuts in Knowledge Editing of
                  Large Language Models},
  booktitle    = {Proceedings of the 62nd Annual Meeting of the Association for Computational
                  Linguistics (Volume 1: Long Papers), {ACL} 2024, Bangkok, Thailand,
                  August 11-16, 2024},
  pages        = {8987--9001},
  publisher    = {Association for Computational Linguistics},
  year         = {2024},
  url          = {https://doi.org/10.18653/v1/2024.acl-long.486},
  doi          = {10.18653/V1/2024.ACL-LONG.486},
  bibsource    = {dblp computer science bibliography, https://dblp.org}
}

@article{DBLP:journals/corr/abs-2404-05405,
  author       = {Zeyuan Allen{-}Zhu and
                  Yuanzhi Li},
  title        = {Physics of Language Models: Part 3.3, Knowledge Capacity Scaling Laws},
  journal      = {CoRR},
  volume       = {abs/2404.05405},
  year         = {2024},
  url          = {https://doi.org/10.48550/arXiv.2404.05405},
  doi          = {10.48550/ARXIV.2404.05405},
  eprinttype    = {arXiv},
  eprint       = {2404.05405},
  bibsource    = {dblp computer science bibliography, https://dblp.org}
}

@article{DBLP:journals/tacl/ElazarKRRHSG21,
  author       = {Yanai Elazar and
                  Nora Kassner and
                  Shauli Ravfogel and
                  Abhilasha Ravichander and
                  Eduard H. Hovy and
                  Hinrich Sch{\"{u}}tze and
                  Yoav Goldberg},
  title        = {Measuring and Improving Consistency in Pretrained Language Models},
  journal      = {Trans. Assoc. Comput. Linguistics},
  volume       = {9},
  pages        = {1012--1031},
  year         = {2021},
  url          = {https://doi.org/10.1162/tacl\_a\_00410},
  doi          = {10.1162/TACL\_A\_00410},
  bibsource    = {dblp computer science bibliography, https://dblp.org}
}

@article{DBLP:journals/corr/abs-2403-00510,
  author       = {Bo Li and
                  Qinghua Zhao and
                  Lijie Wen},
  title        = {{ROME:} Memorization Insights from Text, Probability and Hidden State
                  in Large Language Models},
  journal      = {CoRR},
  volume       = {abs/2403.00510},
  year         = {2024},
  url          = {https://doi.org/10.48550/arXiv.2403.00510},
  doi          = {10.48550/ARXIV.2403.00510},
  eprinttype    = {arXiv},
  eprint       = {2403.00510},
  bibsource    = {dblp computer science bibliography, https://dblp.org}
}

@inproceedings{DBLP:conf/emnlp/WangYXQD00GJX0C24,
  author       = {Mengru Wang and
                  Yunzhi Yao and
                  Ziwen Xu and
                  Shuofei Qiao and
                  Shumin Deng and
                  Peng Wang and
                  Xiang Chen and
                  Jia{-}Chen Gu and
                  Yong Jiang and
                  Pengjun Xie and
                  Fei Huang and
                  Huajun Chen and
                  Ningyu Zhang},
  editor       = {Yaser Al{-}Onaizan and
                  Mohit Bansal and
                  Yun{-}Nung Chen},
  title        = {Knowledge Mechanisms in Large Language Models: {A} Survey and Perspective},
  booktitle    = {Findings of the Association for Computational Linguistics: {EMNLP}
                  2024, Miami, Florida, USA, November 12-16, 2024},
  pages        = {7097--7135},
  publisher    = {Association for Computational Linguistics},
  year         = {2024},
  url          = {https://aclanthology.org/2024.findings-emnlp.416},
  bibsource    = {dblp computer science bibliography, https://dblp.org}
}

@article{DBLP:journals/bstj/Shannon48a,
  author       = {Claude E. Shannon},
  title        = {A mathematical theory of communication},
  journal      = {Bell Syst. Tech. J.},
  volume       = {27},
  number       = {4},
  pages        = {623--656},
  year         = {1948},
  url          = {https://doi.org/10.1002/j.1538-7305.1948.tb00917.x},
  doi          = {10.1002/J.1538-7305.1948.TB00917.X},
  bibsource    = {dblp computer science bibliography, https://dblp.org}
}

@article{DBLP:journals/bstj/Shannon48,
  author       = {Claude E. Shannon},
  title        = {A mathematical theory of communication},
  journal      = {Bell Syst. Tech. J.},
  volume       = {27},
  number       = {3},
  pages        = {379--423},
  year         = {1948},
  url          = {https://doi.org/10.1002/j.1538-7305.1948.tb01338.x},
  doi          = {10.1002/J.1538-7305.1948.TB01338.X},
  bibsource    = {dblp computer science bibliography, https://dblp.org}
}

@article{DBLP:journals/corr/abs-2409-15825,
  author       = {Junjie Ye and
                  Yuming Yang and
                  Qi Zhang and
                  Tao Gui and
                  Xuanjing Huang and
                  Peng Wang and
                  Zhongchao Shi and
                  Jianping Fan},
  title        = {Empirical Insights on Fine-Tuning Large Language Models for Question-Answering},
  journal      = {CoRR},
  volume       = {abs/2409.15825},
  year         = {2024},
  url          = {https://doi.org/10.48550/arXiv.2409.15825},
  doi          = {10.48550/ARXIV.2409.15825},
  eprinttype    = {arXiv},
  eprint       = {2409.15825},
  bibsource    = {dblp computer science bibliography, https://dblp.org}
}

@article{DBLP:journals/corr/abs-2001-08361,
  author       = {Jared Kaplan and
                  Sam McCandlish and
                  Tom Henighan and
                  Tom B. Brown and
                  Benjamin Chess and
                  Rewon Child and
                  Scott Gray and
                  Alec Radford and
                  Jeffrey Wu and
                  Dario Amodei},
  title        = {Scaling Laws for Neural Language Models},
  journal      = {CoRR},
  volume       = {abs/2001.08361},
  year         = {2020},
  url          = {https://arxiv.org/abs/2001.08361},
  eprinttype    = {arXiv},
  eprint       = {2001.08361},
  bibsource    = {dblp computer science bibliography, https://dblp.org}
}


\appendix

\begin{table}[ht]
\centering
\caption{Robustness analysis on low-frequency knowledge. $R^2$ scores are reported for the CO-OCCUR and SMI, restricted to the bottom 20\% of evaluation triples ranked by $P(s,o)$.}
\label{tab:low_frequency_robustness}
\begin{tabular}{lccc}
\toprule
\textbf{Model} & \textbf{$R^2$: CO-OCCUR} & \textbf{$R^2$: SMI} & \textbf{$P(s,o)$ Range} \\ 
\midrule
\multicolumn{4}{l}{\textbf{BLOOM}} \\
bloom-560m     & 0.076 & 0.512 & (0, 1e-02) \\
bloom-1b1      & 0.235 & 0.663 & (0, 1e-02) \\
bloom-1b7      & 0.216 & 0.680 & (0, 1e-02) \\
bloom-3b       & 0.371 & 0.740 & (0, 1e-02) \\
bloom-7b1      & 0.291 & 0.733 & (0, 1e-02) \\
bloom          & 0.202 & 0.573 & (0, 1e-02) \\
\midrule
\multicolumn{4}{l}{\textbf{GPT-NEO}} \\
gpt-neo-125m   & 0.048 & 0.467 & (0, 6e-03) \\
gpt-neo-1.3B   & 0.245 & 0.794 & (0, 6e-03) \\
gpt-neo-2.7B   & 0.192 & 0.771 & (0, 6e-03) \\
gpt-j-6b       & 0.045 & 0.727 & (0, 6e-03) \\
gpt-neox-20b   & 0.022 & 0.693 & (0, 6e-03) \\
\midrule
\multicolumn{4}{l}{\textbf{Pythia}} \\
pythia-14m     & 0.005 & 0.302 & (0, 6e-03) \\
pythia-31m     & 0.020 & 0.316 & (0, 6e-03) \\
pythia-70m     & 0.052 & 0.370 & (0, 6e-03) \\
pythia-160m    & 0.104 & 0.463 & (0, 6e-03) \\
pythia-410m    & 0.111 & 0.643 & (0, 6e-03) \\
pythia-1.4b    & 0.095 & 0.735 & (0, 6e-03) \\
pythia-2.8b    & 0.115 & 0.780 & (0, 6e-03) \\
pythia-6.9b    & 0.097 & 0.785 & (0, 6e-03) \\
pythia-12b     & 0.047 & 0.757 & (0, 6e-03) \\
\midrule
\multicolumn{4}{l}{\textbf{TinyLlama}} \\
TinyLlama-1.1B & 0.108 & 0.726 & (0, 1e-02) \\
\midrule
\multicolumn{4}{l}{\textbf{Ours}} \\
ours-1.6b      & 0.290 & 0.747 & (0, 6e-03) \\
ours-7b        & 0.246 & 0.672 & (0, 6e-03) \\
ours-13b       & 0.075 & 0.601 & (0, 6e-03) \\
\bottomrule
\end{tabular}
\end{table}
\begin{table*}[t]
\caption{Several cases with significant deviations from the co-occurrence metric are shown here. $N_s$ is the number of paragraphs in which the subject appears, $N_o$ is the number of paragraphs in which the object appears, and $N_{s,o}$ is the number of paragraphs in which they co-occur. The first two cases exhibit rare co-occurrence but demonstrate high accuracy, whereas the latter two cases show frequent co-occurrence but exhibit low accuracy.}
\label{case_study}
\begin{center}
\begin{small}
\resizebox{\linewidth}{!}{ 
\begin{tabular}{p{1.7cm} p{2.7cm} p{1.5cm} cccc}
\toprule
Subject & Question & Object & $N_s$ & $N_o$ & $N_{s,o}$ & $ACC$ \\
\midrule
\multicolumn{7}{c}{\text{Rare Co-occurrence but High Accuracy}} \\
\midrule
Csepel SC & The headquarters of Csepel SC is in & Budapest & 398 & 748,860 & 172 & 0.655 \\
Liis Lemsalu & Liis Lemsalu is a citizen of & Estonia & 307 & 967,207 & 160 & 0.570 \\
\midrule
\multicolumn{7}{c}{\text{Frequent Co-occurrence but Low Accuracy}} \\
\midrule
Paris & Paris is located in & Europe & 20,902,571 & 59,408,155 & 3,761,610 & 0.043 \\
Ohio & The capital of Ohio is & Columbus & 11,894,303 & 3,718,483 & 1,210,957 & 0.193 \\
\bottomrule
\end{tabular}
} 
\end{small}
\end{center}
\end{table*}
\begin{table}[ht]
\centering
\caption{Distribution of evaluation samples across accuracy intervals.}
\label{tab:sample_distribution}
\begin{tabular}{lccccc}
\toprule
\textbf{Model} & \textbf{[0, 0.2)} & \textbf{[0.2, 0.4)} & \textbf{[0.4, 0.6)} & \textbf{[0.6, 0.8)} & \textbf{[0.8, 1.0]} \\ 
\midrule
\multicolumn{6}{l}{\textbf{BLOOM}} \\
bloom-560m     & 10108 & 480  & 118  & 75  & 590 \\
bloom-1b1      & 9565  & 763  & 282  & 127 & 634 \\
bloom-1b7      & 9395  & 874  & 300  & 136 & 666 \\
bloom-3b       & 9096  & 991  & 435  & 161 & 688 \\
bloom-7b1      & 8608  & 1170 & 578  & 246 & 769 \\
bloom          & 7474  & 1821 & 1239 & 500 & 337 \\
\midrule
\multicolumn{6}{l}{\textbf{GPT-NEO}} \\
gpt-neo-125m   & 11480 & 318  & 117  & 14  & 600 \\
gpt-neo-1.3B   & 10448 & 881  & 373  & 154 & 673 \\
gpt-neo-2.7B   & 10151 & 1042 & 460  & 199 & 677 \\
gpt-j-6b       & 9012  & 1534 & 813  & 386 & 784 \\
gpt-neox-20b   & 8973  & 1752 & 762  & 317 & 725 \\
\midrule
\multicolumn{6}{l}{\textbf{Pythia}} \\
pythia-14m     & 11924 & 7    & 0    & 0   & 598 \\
pythia-31m     & 11916 & 12   & 3    & 0   & 598 \\
pythia-70m     & 11901 & 29   & 1    & 0   & 598 \\
pythia-160m    & 11687 & 221  & 17   & 6   & 598 \\
pythia-410m    & 11024 & 628  & 197  & 75  & 605 \\
pythia-1.4b    & 10349 & 984  & 378  & 160 & 658 \\
pythia-2.8b    & 9786  & 1292 & 531  & 207 & 713 \\
pythia-6.9b    & 9483  & 1497 & 613  & 200 & 736 \\
pythia-12b     & 9260  & 1579 & 721  & 224 & 745 \\
\midrule
\multicolumn{6}{l}{\textbf{TinyLlama}} \\
TinyLlama-1.1B & 10195 & 1161 & 430  & 143 & 704 \\
\midrule
\multicolumn{6}{l}{\textbf{Ours}} \\
ours-1.6b      & 9929  & 1269 & 511  & 194 & 715 \\
ours-7b        & 7713  & 2138 & 1240 & 618 & 909 \\
ours-13b       & 7707  & 2323 & 1172 & 547 & 869 \\
\bottomrule
\end{tabular}
\end{table}
\begin{table}[t]
\caption{Twenty templates and their corresponding accuracy values for the \(\mathbf{BORN}\) relation on the ours-13b model. {[S]} represents the subject in the knowledge triple.}
\label{templates}
\begin{center}
\begin{tabular}{lc}
\toprule
Template & ACC \\
\midrule
{[S]}’s birth took place in & 0.186 \\
{[S]} was brought into existence in & 0.187 \\
{[S]} was born into the world in & 0.190 \\
{[S]}’s nativity is & 0.197 \\
The place in which {[S]} was given birth to is & 0.201 \\
The hometown of {[S]} is & 0.203 \\
The birthplace of {[S]} is & 0.212 \\
{[S]} was delivered in & 0.215 \\
{[S]}’s cradle was in & 0.219 \\
{[S]}’s entry into the world happened in & 0.225 \\
{[S]} first saw the light in & 0.228 \\
{[S]}’s roots are in & 0.230 \\
{[S]}’s life began in & 0.230 \\
{[S]} was born and raised in & 0.237 \\
{[S]}’s origin is & 0.238 \\
{[S]} entered life in & 0.251 \\
{[S]} came into the world in & 0.255 \\
{[S]} was born in & 0.255 \\
{[S]}’s birth occurred in & 0.263 \\
{[S]}’s starting point was & 0.282 \\
\bottomrule
\end{tabular}
\end{center}
\end{table}

\section{AI Assistants in Research or Writing}

In preparing this manuscript, AI assistants were employed solely to assist with refining the clarity, style, and readability of certain text segments. They were not involved in designing the study, developing or implementing the methodology, collecting or analyzing data, or generating the primary scientific contributions. All substantive research decisions, analyses, and conclusions are fully the responsibility of the authors.

\section{Additional Experimental Results}
\label{appendix:Additional Experimental Results}

\subsection{Validating the Predictive Efficacy and Robustness of SMI}
\label{appendix:low_frequency_robustness}
To verify the reliable predictive power of the proposed Size-dependent Mutual Information (SMI) and to demonstrate that its effectiveness extends beyond simple co-occurrence frequency, we provide additional analyses showing that SMI is not merely a byproduct of subject–object frequency counts in the pre-training data.

The SMI builds on Mutual Information (MI), which depends jointly on joint frequency and marginals. Identical co-occurrence rates can yield vastly different MI values given different marginals. For example, when $P(s,o)=0.1$, $P(s)=0.1$, $P(o)=0.1$, we obtain $I(s,o)=0.23$. With the same $P(s,o)=0.1$ but $P(s)=0.2$, $P(o)=0.4$, the MI drops to $0.02$.
This illustrates that MI, as well as the derived SMI metric, cannot be reduced to co-occurrence frequency alone, as they explicitly account for the specificity induced by the marginal distributions of the subject and object entities.

To empirically validate this distinction, we compare SMI with the pure co-occurrence baseline (CO-OCCUR), which uses only the logarithm of the subject–object co-occurrence count as its predictive signal. As reported in Table 1, SMI consistently and substantially outperforms the CO-OCCUR baseline across all models, demonstrating a marked improvement in $R^2$ scores.

We further evaluate the robustness of SMI by focusing on knowledge triples with low co-occurrence frequency. Specifically, for each model, we sort the evaluation data by $P(s,o)$ and restrict the analysis to the lowest 20\% of examples. We then recompute $R^2$ for both CO-OCCUR and SMI on this specific subset. As summarized in Table~\ref{tab:low_frequency_robustness}, the predictive performance of the CO-OCCUR baseline deteriorates markedly when $P(s,o) < 0.01$; in contrast, SMI sustains a high and stable correlation. This divergence underscores that SMI is not a mere proxy for co-occurrence frequency, but instead captures predictive structure that goes beyond simple frequency priors, particularly in low-exposure regimes.

\subsection{Additional Baselines for Memorization Measures}
\label{appendix:Additional Baselines for Memorization Measures}
Prior studies on knowledge memorization in language models generally identify two dominant factors: the frequency with which knowledge appears in the pre-training data and the model size. Beyond the co-occurrence baseline used in the main experiments, defined as the logarithm of subject–object co-occurrence count ($\log(N \cdot P(s, o))$), we evaluate several alternative metrics that combine these factors in simple ways.

Specifically, we consider variants such as $\log(N \cdot P(s, o)) + \log \Phi$ and $\log(N \cdot P(s, o)) \cdot \log \Phi$. Since these variants are linear transformations of $\log(N \cdot P(s, o))$, they do not change linear correlation results and perform identically to the original co-occurrence baseline.

In addition, we evaluate a baseline that directly uses raw co-occurrence frequency without logarithmic scaling ($N \cdot P(s, o)$). As shown in Table~\ref{tab:raw_frequency_results}, this variant performs substantially worse across models, indicating that the co-occurrence baseline exhibits negligible predictive power if not properly transformed. To the best of our knowledge, there are no stronger baselines in this specific setting.

\begin{table}[ht]
\centering
\caption{Predictive performance ($R^2$) of raw co-occurrence frequency.}
\label{tab:raw_frequency_results}
\begin{tabular}{lc}
\toprule
\textbf{Model} & \textbf{$R^2$: Raw Co-occurrence Frequency} \\ 
\midrule
\multicolumn{2}{l}{\textbf{BLOOM}} \\
bloom-560m     & 0.002 \\
bloom-1b1      & 0.001 \\
bloom-1b7      & 0.002 \\
bloom-3b       & 0.000 \\
bloom-7b1      & 0.003 \\
bloom          & 0.016 \\
\midrule
\multicolumn{2}{l}{\textbf{GPT-NEO}} \\
gpt-neo-125m   & 0.020 \\
gpt-neo-1.3B   & 0.004 \\
gpt-neo-2.7B   & 0.005 \\
gpt-j-6b       & 0.000 \\
gpt-neox-20b   & 0.000 \\
\midrule
\multicolumn{2}{l}{\textbf{Pythia}} \\
pythia-14m     & 0.039 \\
pythia-31m     & 0.037 \\
pythia-70m     & 0.062 \\
pythia-160m    & 0.022 \\
pythia-410m    & 0.000 \\
pythia-1.4b    & 0.000 \\
pythia-2.8b    & 0.001 \\
pythia-6.9b    & 0.000 \\
pythia-12b     & 0.000 \\
\midrule
\multicolumn{2}{l}{\textbf{TinyLlama}} \\
TinyLlama-1.1B & 0.000 \\
\midrule
\multicolumn{2}{l}{\textbf{Ours}} \\
ours-1.6b      & 0.023 \\
ours-7b        & 0.024 \\
ours-13b       & 0.010 \\
\bottomrule
\end{tabular}
\end{table}

\subsection{Statistical Comparison Between MI and SMI}
To assess whether the performance difference between MI and SMI observed in Table~\ref{main_table} is statistically significant, we conduct a paired-sample $t$-test on their $R^2$ values across the 24 evaluated models.

The analysis yields a test statistic of $t(23) = 20.00$ with a corresponding $p$-value of 4.84e-16. The 95\% confidence interval is $[0.003, 0.004]$. Since the $p$-value is substantially below the conventional significance threshold ($0.05$), this result indicates that the improvement of SMI over MI is statistically significant.

While the absolute difference in $R^2$ is modest, SMI consistently outperforms MI across all evaluated models. Moreover, SMI is computed directly from MI with negligible additional computational overhead. Collectively, these properties demonstrate that SMI provides a reliable and practically valuable refinement over MI.

\subsection{Analysis in High-SMI Regions}
\label{appendix:Analysis in High-SMI Regions}
In Figure~\ref{smi_figures}, an increased dispersion in the linear correlation between accuracy and SMI is observed within high-SMI regions. This behavior is primarily attributable to the evaluation design and the intrinsic difficulty of the task, rather than a limitation of SMI itself.

For each question type, we evaluate models using 20 paraphrased prompts. Although these prompts are semantically equivalent (see Appendix~\ref{appendix:Templates for Knowledge Triples}), some formulations are relatively uncommon in the pre-training corpus and therefore more challenging for language models. As a result, even large-scale models do not consistently achieve high accuracy across all prompts. As shown in Figure~\ref{pythia_acc}, BLOOM-176B attains an average accuracy of only approximately 19\%, indicating that high absolute accuracy is not a necessary indicator of strong model performance under this evaluation setting.

In addition, the number of samples achieving high accuracy is limited. This sparsity leads to increased statistical variance in the high-accuracy regime, which in turn results in greater dispersion of data points when SMI values are high. Table~\ref{tab:sample_distribution} reports the number of samples per model within each accuracy interval. Notably, the majority of samples fall below an accuracy of 0.4, which helps explain the increased dispersion observed in high-SMI regions.

\section{Licenses and Terms of Use}

We use the following publicly available assets in this work:

\begin{itemize}
    \item \textbf{ParaRel:} \url{https://github.com/yanaiela/pararel}, MIT License~\citep{DBLP:journals/tacl/ElazarKRRHSG21}.
    \item \textbf{The Pile:} \url{https://huggingface.co/datasets/EleutherAI/pile}, MIT License~\citep{DBLP:journals/corr/abs-2101-00027}.
    \item \textbf{ROOTS (En):} \url{https://huggingface.co/bigscience-data}, cc-by-nc-sa-4.0 license~\citep{DBLP:conf/nips/LaurenconSWAMSW22}.
    \item \textbf{SlimPajama:} \url{https://huggingface.co/datasets/cerebras/SlimPajama-627B}, Common Crawl Foundation Terms of Use, C4 license, MIT License, BSD License, Apache License, the\_pile\_books3 license, pg19 license, ArXiv Terms of Use, Wikipedia License, StackExchange license on the Internet Archive~\citep{cerebras2023slimpajama}.
    \item \textbf{RefinedWeb:} \url{https://huggingface.co/datasets/tiiuae/falcon-refinedweb}, ODC-By 1.0 license, CommonCrawl ToU~\citep{DBLP:conf/nips/PenedoMHCACPAL23}.
    \item \textbf{Wikipedia:} \url{https://huggingface.co/datasets/wikimedia/wikipedia}, GNU Free Documentation License, Creative Commons Attribution-Share-Alike 3.0 License~\citep{wikidump}.
    \item \textbf{bloom-560m:} \url{bigscience/bloom-560m}, RAIL License v1.0~\citep{DBLP:journals/corr/abs-2211-05100}.
    \item \textbf{bloom-1b1:} \url{bigscience/bloom-1b1}, RAIL License v1.0~\citep{DBLP:journals/corr/abs-2211-05100}.
    \item \textbf{bloom-1b7:} \url{bigscience/bloom-1b7}, RAIL License v1.0~\citep{DBLP:journals/corr/abs-2211-05100}.
    \item \textbf{bloom-3b:} \url{bigscience/bloom-3b}, RAIL License v1.0~\citep{DBLP:journals/corr/abs-2211-05100}.
    \item \textbf{bloom-7b1:} \url{bigscience/bloom-7b1}, RAIL License v1.0~\citep{DBLP:journals/corr/abs-2211-05100}.
    \item \textbf{bloom:} \url{bigscience/bloom}, RAIL License v1.0~\citep{DBLP:journals/corr/abs-2211-05100}.
    \item \textbf{gpt-neo-125m:} \url{https://huggingface.co/EleutherAI/gpt-neo-125m}, MIT License~\citep{gpt-neo}.
    \item \textbf{gpt-neo-1.3B:} \url{https://huggingface.co/EleutherAI/gpt-neo-1.3B}, MIT License~\citep{gpt-neo}.
    \item \textbf{gpt-neo-2.7B:} \url{https://huggingface.co/EleutherAI/gpt-neo-2.7B}, MIT License~\citep{gpt-neo}.
    \item \textbf{gpt-j-6b:} \url{https://huggingface.co/EleutherAI/gpt-j-6b}, Apache 2.0 license~\citep{mesh-transformer-jax}.
    \item \textbf{gpt-neox-20b:} \url{https://huggingface.co/EleutherAI/gpt-neox-20b}, Apache 2.0 license~\citep{DBLP:journals/corr/abs-2204-06745}.
    \item \textbf{pythia-14m:} \url{https://huggingface.co/EleutherAI/pythia-14m}, Apache 2.0 license~\citep{DBLP:conf/icml/BidermanSABOHKP23}.
    \item \textbf{pythia-31m:} \url{https://huggingface.co/EleutherAI/pythia-31m}, Apache 2.0 license~\citep{DBLP:conf/icml/BidermanSABOHKP23}.
    \item \textbf{pythia-70m:} \url{https://huggingface.co/EleutherAI/pythia-70m}, Apache 2.0 license~\citep{DBLP:conf/icml/BidermanSABOHKP23}.
    \item \textbf{pythia-160m:} \url{https://huggingface.co/EleutherAI/pythia-160m}, Apache 2.0 license~\citep{DBLP:conf/icml/BidermanSABOHKP23}.
    \item \textbf{pythia-410m:} \url{https://huggingface.co/EleutherAI/pythia-410m}, Apache 2.0 license~\citep{DBLP:conf/icml/BidermanSABOHKP23}.
    \item \textbf{pythia-1.4b:} \url{https://huggingface.co/EleutherAI/pythia-1.4b}, Apache 2.0 license~\citep{DBLP:conf/icml/BidermanSABOHKP23}.
    \item \textbf{pythia-2.8b:} \url{https://huggingface.co/EleutherAI/pythia-2.8b}, Apache 2.0 license~\citep{DBLP:conf/icml/BidermanSABOHKP23}.
    \item \textbf{pythia-6.9b:} \url{https://huggingface.co/EleutherAI/pythia-6.9b}, Apache 2.0 license~\citep{DBLP:conf/icml/BidermanSABOHKP23}.
    \item \textbf{pythia-12b:} \url{https://huggingface.co/EleutherAI/pythia-12b}, Apache 2.0 license~\citep{DBLP:conf/icml/BidermanSABOHKP23}.
    \item \textbf{TinyLlama:} \url{https://huggingface.co/TinyLlama/TinyLlama-1.1B-intermediate-step-480k-1T}, Apache 2.0 license~\citep{DBLP:journals/corr/abs-2401-02385}.
\end{itemize}

We confirm that all licenses and terms of use are respected.

\section{Case Study}
Table~\ref{case_study} highlights four counterintuitive cases in our pre-training data. From these, we observe the following: (1) In the first two cases, the subject and object rarely co-occur, suggesting a seemingly weak relationship. However, whenever the subject does appear, the object almost always appears alongside it, resulting in high accuracy. (2) In the last two cases, despite frequent co-occurrence between the subject and object, their individual occurrence frequencies are significantly higher, leading to low accuracy. For example, when prompted with ``Paris is located in'', the model is more likely to respond with ``France'' rather than ``Europe''.

These examples highlight the pivotal role of knowledge specificity in the memory of LLMs. It is not simply the frequencies of knowledge occurrence that enhance retention, but rather the specificity of the knowledge that proves more influential. This insight suggests that in downstream task training, even when domain-specific data is significantly fewer than the pre-training data, minimizing repetition with the pre-training data can increase the SMI, making it easier for LLMs to retain the relevant knowledge.

\section{Templates for Knowledge Triples}
\label{appendix:Templates for Knowledge Triples}
We create twenty templates for each type of knowledge triple. Table~\ref{templates} illustrates the templates for the \(\mathbf{BORN}\) relation, along with their corresponding evaluation accuracy on the ours-13b model. As shown, the choice of template has a significant impact on the model’s accuracy in answering questions. To address the instability of evaluations based on a single template, we calculate the average accuracy across all 20 templates.
We employ \textbf{gpt-4o-2024-08-06} to generate query templates. The prompt we use is: \textbf{Generate 20 diverse and fluent paraphrases of the following sentence, preserving its original meaning:\textbackslash n[sentence]}. Our evaluation set contains a total of 12,671 pieces of data and 300 query templates, following the MIT license. The template designs for each type of knowledge triple are as follows:
\begin{Verbatim}[breaklines]
{
    "plays": [
        "[S] plays",
        "[S] performs",
        "[S] engages in playing",
        "[S] is an musician of",
        "[S] is involved in playing",
        "[S] is a performer of",
        "[S] executes",
        "[S] is a practitioner of",
        "[S] is an player of",
        "[S] is an artist of",
        "The genre that [S] performs is",
        "[S] captivates audiences by performing",
        "[S] performs the musical style of",
        "[S] is known for his performances in",
        "[S]'s musical expertise lies in",
        "[S] is recognized for playing",
        "[S] is a notable performer of",
        "[S]'s artistry is showcased in",
        "[S]'s musical inclinations are towards",
        "The musical category that [S] excels in is"
    ],
    "occupation": [
        "The occupation of [S] is",
        "[S] works as an",
        "[S] is employed in the profession of an",
        "[S] pursues a profession as an",
        "[S] is a professional",
        "[S] is engaged in the profession of an",
        "[S]'s vocation is an",
        "[S] is an professional",
        "[S]'s role in life is that of an",
        "[S]'s professional identity is that of an",
        "The job of [S] is",
        "[S] is employed as an",
        "[S] has a career as an",
        "The profession of [S] is",
        "[S]'s line of work is",
        "[S] makes her career as an",
        "The career of [S] is",
        "[S] has chosen to be an",
        "[S] earns a living as an",
        "[S]'s professional life revolves around being an"
    ],
    "work": [
        "[S] works in the field of",
        "[S] specializes in the field of",
        "[S]'s work focuses on",
        "[S] is engaged in the area of",
        "[S]'s professional activity lies in",
        "[S]'s expertise is in",
        "[S] operates within the field of",
        "[S]'s research is concentrated in",
        "[S] works extensively in",
        "[S] has dedicated his career to",
        "[S] is involved in the study of",
        "[S]'s field of study is",
        "[S] contributes to the discipline of",
        "[S] practices in the area of",
        "[S]'s work domain is",
        "[S] conducts his work in",
        "[S]'s field of expertise is",
        "[S]'s career revolves around",
        "[S] works professionally in",
        "[S]'s occupation relates to"
    ],
    "is-located": [
        "[S] is located in",
        "The fatherland of [S] is in",
        "[S] can be found in",
        "[S] lives in",
        "[S] is situated within",
        "[S] is based in",
        "[S] resides in",
        "The country in which [S] lives is",
        "[S]'s location is in",
        "[S] stays in",
        "[S]'s address is in",
        "[S] is present in",
        "[S] spends most of his time in",
        "[S] belongs to",
        "[S]'s position is in",
        "You can find [S] in",
        "[S] is positioned in",
        "[S] exists in",
        "[S] stays primarily in",
        "The country in which [S] is located is"
    ],
    "locate": [
        "[S] is located in",
        "The place in which [S] is held is",
        "[S] is held in",
        "The place where [S] takes place is",
        "[S] takes place in",
        "[S] is hosted in",
        "The venue for [S] is",
        "[S] happens in",
        "[S] is organized in",
        "[S] is staged in",
        "[S] occurs in",
        "The place for [S] is",
        "The location of [S] is",
        "[S] is based in",
        "[S] is set in",
        "[S]'s venue is in",
        "[S] is arranged in",
        "[S] is situated in",
        "People can take part in [S] in",
        "[S]'s setting is in"
    ],
    "found": [
        "[S] was founded in",
        "[S] originated in",
        "[S] was established in",
        "[S] came into existence in",
        "[S] was formed in",
        "[S] had its beginnings in",
        "[S] started in",
        "[S] was created in",
        "[S] was initiated in",
        "[S] took shape in",
        "[S] was launched in",
        "[S] began in",
        "[S] emerged in",
        "The foundation of [S] took place in",
        "[S] originated from",
        "[S] was set up in",
        "[S] was first formed in",
        "[S] came to be in",
        "[S] had its origin in",
        "[S]'s formation occurred in"
    ],
    "born": [
        "[S] was born in",
        "[S] came into the world in",
        "The birthplace of [S] is",
        "[S] entered life in",
        "[S] was delivered in",
        "[S]'s origin is",
        "[S] first saw the light in",
        "[S]'s birth occurred in",
        "[S] was brought into existence in",
        "[S]'s nativity is",
        "The place in which [S] was given birth to is",
        "[S] was born and raised in",
        "[S]'s roots are in",
        "The hometown of [S] is",
        "[S]'s cradle was in",
        "[S]'s entry into the world happened in",
        "[S]'s birth took place in",
        "[S] was born into the world in",
        "[S]'s life began in",
        "[S]'s starting point was"
    ],
    "die": [
        "[S] died in",
        "[S] passed away in",
        "[S]'s death occurred in",
        "[S] lost his life in",
        "[S]'s demise happened in",
        "[S] perished in",
        "[S]'s end came in",
        "[S]'s life ended in",
        "[S] met his end in",
        "[S]'s passing was in",
        "[S]'s death took place in",
        "[S]'s final moments were in",
        "[S]'s last breath was in",
        "[S]'s life was lost in",
        "[S]'s expiration occurred in",
        "[S]'s fatality was in",
        "[S]'s death scene was in",
        "The place in which [S] passed away is",
        "[S]'s death location is",
        "[S]'s death site is"
    ],
    "air": [
        "[S] was originally aired on",
        "[S] first aired on",
        "[S] premiered on",
        "[S] was initially broadcast on",
        "[S] debuted on",
        "[S] was first shown on",
        "[S] was originally broadcast on",
        "[S] was first transmitted on",
        "[S] was first telecast on",
        "[S] was first screened on",
        "[S] was first presented on",
        "[S] had its first airing on",
        "[S] was first televised on",
        "[S] was first released on",
        "[S] was first featured on",
        "[S] was first displayed on",
        "[S] was first exhibited on",
        "[S] was first put on air on",
        "[S] was first launched on",
        "[S] was first introduced on"
    ],
    "headquarter": [
        "The headquarters of [S] is in",
        "[S]'s headquarters is in",
        "[S]'s main office is in",
        "[S]'s central office is in",
        "[S]'s head office is in",
        "The corporate office of [S] is in",
        "[S]'s principal office is in",
        "[S]'s administrative center is in",
        "[S]'s base of operations is in",
        "[S]'s HQ is in",
        "The headquarters's location of [S] is in",
        "[S]'s headquarters office is in",
        "[S]'s primary office is in",
        "[S]'s main administrative office is in",
        "[S]'s central headquarters is in",
        "[S]'s main base is in",
        "The main hub of [S] is in",
        "[S]'s main building is in",
        "[S]'s headquarters is located in",
        "The place in which [S]'s headquarters is situated in"
    ],
    "capital": [
        "[S] is the capital of",
        "[S] serves as the capital of",
        "[S] is recognized as the capital of",
        "[S] stands as the capital of",
        "[S] is known as the capital of",
        "[S] functions as the capital of",
        "[S] is officially the capital of",
        "[S] holds the title of capital of",
        "[S] is designated as the capital of",
        "[S] is the administrative capital of",
        "[S] is the political capital of",
        "[S] is the governmental capital of",
        "[S] is the principal capital of",
        "[S] is the main capital of",
        "[S] is the chief capital of",
        "[S] is the primary capital of",
        "[S] is the foremost capital of",
        "The country whose capital is [S] is",
        "[S] is known to the people as the capital of",
        "As the capital city, [S] has an extremely important place in"
    ],
    "citizen": [
        "[S] is a citizen of",
        "The country where [S] holds citizenship and enjoys benefits is",
        "The nation that grants [S] the rights and privileges of citizenship is",
        "[S] is from the country of",
        "The place of [S]'s birth and upbringing is",
        "[S] belongs to",
        "[S]'s homeland is",
        "[S]'s home country is",
        "[S] is a resident of",
        "[S]'s origin is",
        "The land where [S]'s roots run deep is",
        "[S]'s place of origin is",
        "[S]'s country of origin is",
        "The nation that [S] calls home is",
        "[S]'s native country is",
        "[S]'s nationality is from",
        "The place where [S] was born and raised is",
        "[S]'s citizenship gives him the rights and privileges of a citizen in",
        "[S]'s citizenship makes him a part of the country's community in",
        "The country that [S] resides in and holds citizenship is"
    ],
    "create": [
        "[S] was created in",
        "The birthplace of [S] is",
        "[S] originated in",
        "The invention of [S] took place in",
        "[S] hails from",
        "The country that [S] was made in is",
        "[S] was initially developed in",
        "The country that introduced [S] is",
        "[S] was crafted in",
        "The place of origin for [S] is",
        "[S] was first produced in",
        "The homeland of [S] is",
        "The source of [S] is",
        "[S] was manufactured in",
        "The origin of [S]'s creation is",
        "[S] was designed in",
        "The place where [S] was conceived is",
        "[S] was made in",
        "The birth nation of [S] is",
        "[S] was developed in"
    ],
    "play": [
        "[S] plays",
        "[S] is known for playing the",
        "[S]'s instrument is the",
        "[S] is a musician who plays the",
        "[S] is one who has mastered the art of the",
        "[S] has a gift for playing the",
        "Few can rival [S]'s prowess on the",
        "[S] is a true virtuoso of the",
        "[S]'s music is infused with the soulful sound of the",
        "Audiences are captivated by [S]'s",
        "[S]'s musical talents shine through his",
        "[S] is a skilled",
        "One can hear [S]'s passion in every",
        "With every note, [S] demonstrates his mastery of the",
        "[S] is a",
        "[S] plays the",
        "[S] specializes in the",
        "[S] performs on the",
        "Music flows through [S] and his",
        "As for instruments, [S] prefers the"
    ],
    "is-locate": [
        "[S] is located in",
        "[S] lies in the",
        "[S] lies within the boundaries of",
        "The place where you can find [S] is",
        "[S] is situated in",
        "[S] is positioned in",
        "[S] exists in",
        "[S] is placed in",
        "[S] resides in",
        "The place where [S] can be fully appreciated is",
        "[S] can be found in",
        "[S] is in the area of",
        "[S] is part of the landscape of",
        "[S] is in the lands of",
        "The region that holds [S] is",
        "The location of [S] is",
        "[S] is in the sector of",
        "The place that [S] is a part of is",
        "[S] is in the section of",
        "[S] is in the region of"
    ]
}

\end{Verbatim}

\end{document}